\documentclass[lettersize,journal]{IEEEtran}
\usepackage{cite}
\usepackage{amsmath,amssymb,amsfonts}
\usepackage{textcomp}
\usepackage{wrapfig}
\usepackage{graphicx, subcaption}
\usepackage{multirow}
\usepackage{multicol}
\usepackage{makecell}
\usepackage{xcolor}
\usepackage{subfiles}
\usepackage{algorithm2e}

\graphicspath{{images/}}

\begin{document}
\title{
Optimising complexity of CNN models for resource constrained devices: QRS detection case study
}

\author{Ahsan Habib, \IEEEmembership{Graduate Student Member, IEEE}, Chandan Karmakar, and John Yearwood, \IEEEmembership{Member, IEEE}
\thanks{This paragraph of the first footnote will contain the date on 
which you submitted your paper for review.}
\thanks{Ahsan Habib, John Yearwood and Chandan
Karmakar are with the School of Information Technology, Deakin University, Geelong, 3225, Australia (e-mail: mahabib@deakin.edu.au, karmakar@deakin.edu.au, john.yearwood@deakin.edu.au).}
}

\markboth{Journal of \LaTeX\ Class Files,~Vol.~14, No.~8, August~2021}%
{Shell \MakeLowercase{\textit{et al.}}: A Sample Article Using IEEEtran.cls for IEEE Journals}

\IEEEpubid{0000--0000/00\$00.00~\copyright~2021 IEEE}

\maketitle

\begin{abstract}

Traditional DL models are complex and resource hungry and thus, care needs to be taken in designing Internet of (medical) things (IoT, or IoMT) applications balancing efficiency-complexity trade-off. 
Recent IoT solutions tend to avoid using deep-learning methods due to such complexities, and rather classical filter-based methods are commonly used.
We hypothesize that a shallow CNN model can offer satisfactory level of performance in combination by leveraging other essential solution-components, such as post-processing that is suitable for resource constrained environment.
In an IoMT application context, QRS-detection and R-peak localisation from ECG signal as a case study, the complexities of CNN models and post-processing were varied to identify a set of combinations suitable for a range of target resource-limited environments.
To the best of our knowledge, finding a deploy-able configuration, by incrementally increasing the CNN model complexity, as required to match the target's resource capacity, and leveraging the strength of post-processing, is the first of its kind.
The results show that a shallow 2-layer CNN with a suitable post-processing can achieve $>$90\% F1-score, and the scores continue to improving for 8-32 layer CNNs, which can be used to profile target constraint environment.
The outcome shows that it is possible to design an optimal DL solution with known target performance characteristics and resource (computing capacity, and memory) constraints. 

\end{abstract}


\section{Introduction} \label{sec:introduction}

\IEEEPARstart{O}{ften}, deep-learning (DL) based solutions stress to design a model complex enough, in terms of depth or other techniques, to understand a phenomena and achieve high performance for a given task.
This is particularly valid for computer vision tasks~\cite{ronneberger2015u,simonyan2014very,szegedy2015going}, as well as, physiological time-series tasks~\cite{hannun_cardiologist-level_2019,cai_qrs_2020}.
A deep model consists of a large number of layers, thus, a large number of parameters which require high computation capacity to train with a big enough dataset.
The time and space complexity of such a deep model to deploy often exceeds the capacity of resource constrained environments (computing capacity, memory, energy etc.).
The use of DL-based models in the client-side environment (sensor, smart-watch, or smart-phone) of IoT-based solutions, particularly, Internet of Medical Things (IoMT) based physiological signal monitoring solution are still unpopular and traditional filter-based approaches are commonly used~\cite{zhao2021robust,chen2017qrs,tang2018real,berwal2018design}.
A shallow DL model may not be outstanding but has the capacity to offer a better solution together by leveraging the strengths of other essential components in an end-to-end IoMT solution architecture, which still lacks a rigorous exploration.

\begin{figure*}[t]
  \centering
  \begin{subfigure}[b]{0.87\textwidth}
    \includegraphics[width=\textwidth]{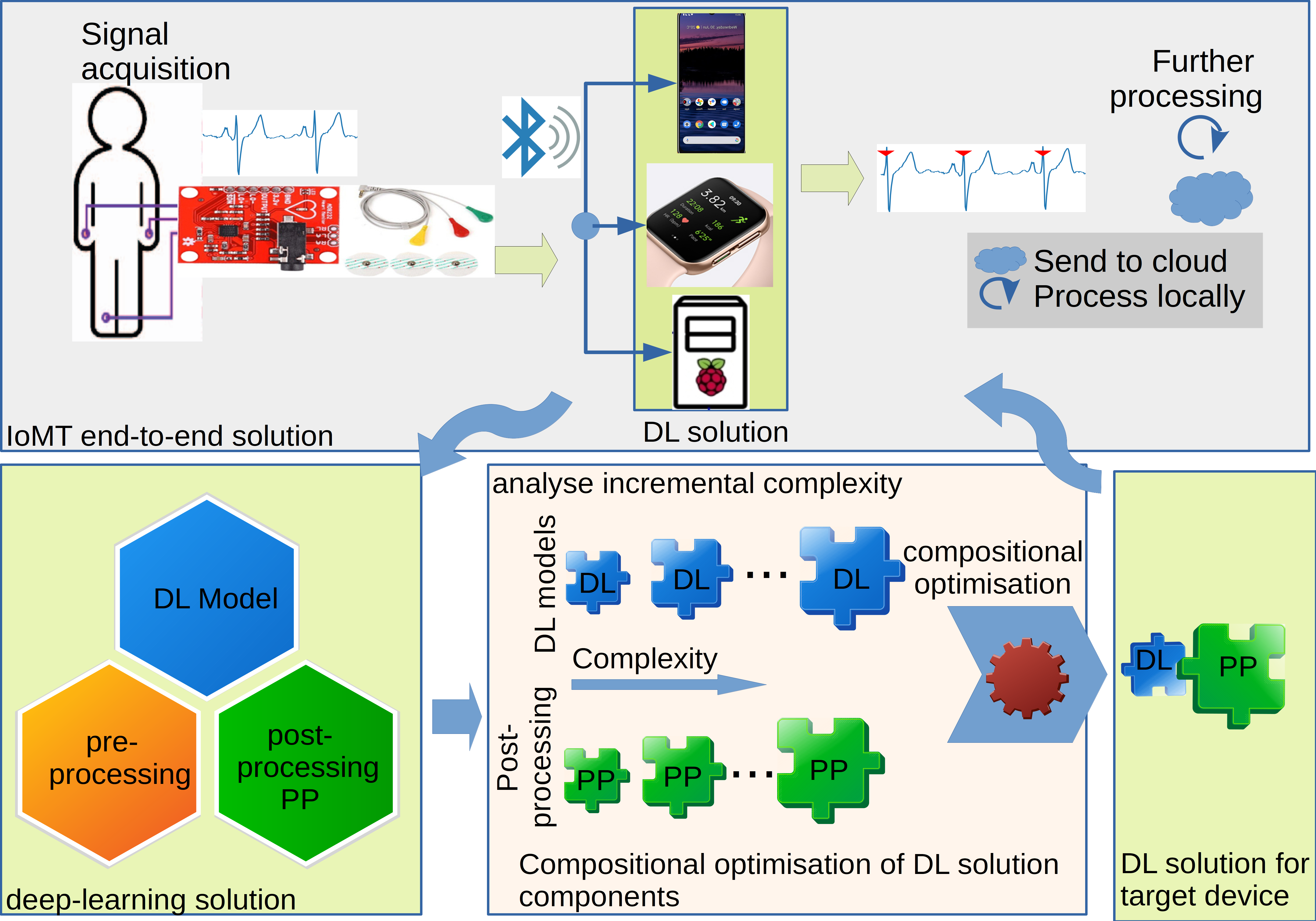}
  \end{subfigure}
  \caption{A conceptual block-diagram showing a deep-learning solution's decomposition to its constituent parts, including pre-processing, DL model, and post-processing. Pre-processing mainly used to prepare an environment, so the remaining two parts were analysed varying their complexities in order to find a suitable compositional solution for a target resource constrained environment.}
  \label{fig:cnn_pp_config}
\end{figure*}

DL algorithms, as well as, filter-based classical methods which deal with physiological signals often require the input signal to pre-process and the algorithm's output to post-process before a decision is finalised.
The back-propagation training for DL models generally require the data to contain a certain-level of noise, in case where the data is clean, noise from external source is added to it so that the model is forced to be robust enough to learn sufficient characteristic features and be able to reduce over-fitting and better generalise over unknown test data~\cite{reed1999neural,bishop1995neural,chen1994robust}.
This characteristic of noise robustness, in particular, is crucial for tasks with physiological signal which inherently contaminated with noise from multiple sources where threshold-based classical methods commonly fail to generalise over unknown test data.
Classical filter-based methods in QRS-detection literature, for example, rarely report performance with multiple ECG datasets~\cite{ahlstrom1983automated,baharestani1979heart,borjesson1982adaptive,afonso1999ecg,bahoura1997dsp}, whereas, recent DL approaches were found to test models across a broad range of datasets~\cite{cai_qrs_2020,Xiang2018}.
In spite of this benefit, recent IoMT studies often found to using traditional methods~\cite{ahlstrom1983automated,baharestani1979heart,borjesson1982adaptive,afonso1999ecg,bahoura1997dsp}.
This reluctance to using DL models can probably be characterised by the resource required by deep models which seem to be overwhelming for resource constrained environment.

A solution to a given physiological bio-signal related task, offered by a DL method, can be decomposed to its components including a pre-processing, the model itself, and a post-processing, as shown in Figure~\ref{fig:cnn_pp_config}.
Since the pre-processing is mainly used to prepare an environment, remaining two components were considered as variable.
This study decouples the CNN models from post-processing to clearly understand their relative importance, later their complexities were varied to leverage the strengths of their combinations in order to find a solution customised for a resource constrained environment.
Considering a DL-based solution as a compositional optimisation of its components, that is suitable for resource constrained environment, were not actively sought in recent IoMT studies, which could be an attractive alternative to traditional approaches currently used.

A task of QRS-detection and R-peak localisation was selected as a case study to explore the strategy of incrementally increasing the complexity of a DL-based method's components (DL model and post-processing) to offer an IoMT solution for continuous ECG signal monitoring.
Monitoring physiological signals using wearable sensors is increasing in recent time and monitoring ECG signal is of particular importance since it can detect a variety of heart problems, including arrhythmias, coronary heart disease, heart attacks, and cardiomyopathy~\cite{de1998prognostic,valles2010ecg,luz2016ecg,guo_inter-patient_2019}.
ECG signal consists of P-wave, QRS-complex, and T-wave. 
The QRS-detection literature contains traditional digital filter-based approaches~\cite{Kohler2002,ahlstrom1983automated,baharestani1979heart,borjesson1982adaptive,afonso1999ecg,bahoura1997dsp} and classical machine-learning approaches~\cite{mehta2008svm,kropf2017ecg,saini2013qrs} to detect QRS and localise R-peaks.
Recent deep-learning-based QRS-detection and R-peak localisation studies reported superior performance across multiple validation-datasets, which involves the use of complex CNN-model-architectures and essential post-processing \cite{cai_qrs_2020,jia_high_2019,liu_octave_2020,sarlija_convolutional_2017,Lee2018,LEE201966}.
The post-processing (PP) in these studies can be summarised as -
\begin{itemize}
    \item minimal PP (basic Salt and Pepper filter)~\cite{jia_high_2019},
    \item moderate PP (use domain-knowledge to group 1s as QRS candidate, i.e. 64ms equivalent sample-length)~\cite{cai_qrs_2020, Lee2018, LEE201966, sarlija_convolutional_2017}, and
    \item advanced PP (use domain-knowledge to filter close QRS neighbours, i.e. minimum R-R distance to be 100 milliseconds)~\cite{jia_high_2019, LEE201966,cai_qrs_2020}
\end{itemize}
Sometime it is unclear if the resultant QRS-detection performance is due to the strength of the model or the post-processing and very limited information is available to quantify the post-processing effect.
In addition, none of these deep learning based approaches consider limitations of resource constrained devices for their algorithm design.

This study identifies relative complexities of CNN models and post-processing so that a set of combinations can be obtained.
CNN model's complexity was increased by increasing its depth, since the depth parameter significantly contributes to a model's learning capacity~\cite{simonyan2014very} and the complexity of post-processing was dealt by segregating to three components based on the amount of domain-knowledge used.
A set of model-and-postprocessing configurations was then identified to be suitable for corresponding target environments' resource capacity. 
To the best of our knowledge, the selection of a CNN model by incrementally increasing its complexity, based on a target resource-constraint environment, and leveraging the strength of post-processing to propose a set of configurations is unique of its kind.

The implementation is released as open source on GitHub.\footnote{https://github.com/deakin-deep-dreamer/qrs\_postprocess012}

\section{METHODOLOGY} \label{sec:method}

\subsection{Problem Formulation}
The QRS-detection and R-peak localisation task was formulated as a segmentation problem where each input sample receives a prediction through a CNN model.
By labeling five samples around an annotated R-location (approx. 0.05 seconds per side, forming a 0.1 second QRS region) as 1s and the rest samples as 0s, similar to Jia et al. ~\cite{jia_high_2019}, a binary-mask was created to train a CNN model to be able to predict a similar mask which later used to localise R-peaks.

\subsection{Application Framework} \label{app_framework}
The proposed QRS detection algorithm fits within an application framework as shown in Figure~\ref{fig:cnn_pp_config}.
ECG signal can be acquired using sensors and sent to a destination using wireless protocols, including Bluetooth low-energy (BLE), among others.
A resource constraint embedded device works as an end-point to receive raw ECG signal.
A suitable QRS detection method can be determined by leveraging the strengths of DL model and required post-processing for a target resource-constraint environment, including smart-phone, smart-watch, or an embedded router hardware, which may connect multiple sensors Figure~\ref{fig:cnn_pp_config}).
The output information (ECG signal and R-peak locations) can be processed locally or could be transferred to the cloud server for further processing.

\subsection{ECG Data}
ECG datasets from PhysioNet data-bank \cite{goldberger_physiobank_2000} were used for this study (summarised in Table \ref{tab:tbl_databases}), 
\begin{table}[h!]
    \centering
    \caption{Characteristics of PhysioNet datasets used in current study.}
    \begin{tabular}{cccccc}
      \hline
      \emph{DB-Name} & \emph{\makecell{Source\\Hz}} & \emph{\makecell{No. of\\Rec.}} & \emph{\makecell{Used no.\\of Rec.}} & \emph{\makecell{Length\\(minute/rec.)}} & \emph{\makecell{No.\\of Beats}} \\
      \hline
      EDB & 250 & 90 & 90 & 120 & 791665 \\
      INCART & 257 & 75 & 75 & 30 & 175900 \\
      MIT-BIH-Arr & 360 & 48 & 46 & 30 & 102941 \\
      NSTDB & 360 & 12 & 12 & 30 & 25590 \\
      QT & 250 & 82 & 80 & 15 & 84883 \\
      SVDB & 128 & 78 & 78 & 30 & 184583 \\
      STDB & 360 & 28 & 28 & vary & 76175 \\
      TWADB & 500 & 100 & 100 & 2 & 18993 \\
      \hline
    \end{tabular}
    \label{tab:tbl_databases}
\end{table}
including the MIT-BIH-Arrhythmia \cite{moody_impact_2001}, INCART, QT \cite{laguna_database_1997}, EDB (European ST-T Database) \cite{taddei_european_1992}, STDB (MIT-BIH ST Change Database), TWADB (T-Wave Alternans Challenge Database), NSTDB (MIT-BIH Noise Stress Test Database) \cite{moody_noise_1984}, and SVDB (MIT-BIH Supraventricular Arrhythmia Database) \cite{greenwald1990improved}.
The first channel signal, among others, was used.
The STDB contains variable length records and on average, 28 minutes long.

\subsection{Pre-processing}
\paragraph{Resampling} 
PhysioNet datasets are sampled at a range of frequencies, so they were resampled at a minimal 100Hz frequency, considering the fact that average QRS-complex may go up to 25Hz or beyond \cite{ajdaraga_analysis_2017}, so the Nyquest sampling frequency \cite{proakis_digital_2004} should be at least 50Hz.

\paragraph{Segmentation} 
A three second segmentation was applied with two seconds overlapping, allowing each QRS region to receive the model prediction multiple times, which may increase the detection likelihood.
A three second segmentation was applied with two seconds overlapping, allowing more long-enough segments for the training, without increasing the model's complexity significantly (longer input segment increases convolution operation computation).
For test ECG records, a non-overlapped 3s segmentation was used to generate predictions. 

\paragraph{Normalisation} The heterogeneous datasets were normalised using standard-scores and such normalisation was performed at segment-level instead of the recording-level to localise the effect of noise.

\subsection{CNN Model}
The focus of this study was to observe the performance of CNN models, with a gradual increase of their complexity (by increasing their depths~\cite{simonyan2014very}, in the range of 2-64 layers) and leverage the strength of post-processing (PP 1-3) to yield a set of configurations for a range of resource-constrained environments.
Shallow CNN models are often utilised in QRS detection~\cite{chandra_robust_2019,Xiang2018,sarlija_convolutional_2017}, thus a simple convolution-only model (no sub-sampling layer, with output dimension the same as the input, using a similar philosophy of Pelt et al.~\cite{pelt_mixed-scale_2018}) was selected as a baseline model, which will be referred as baseline-convnet in the text.
\begin{figure}[h]
  \centering
  \begin{subfigure}[b]{0.29\textwidth}
    \includegraphics[width=\textwidth]{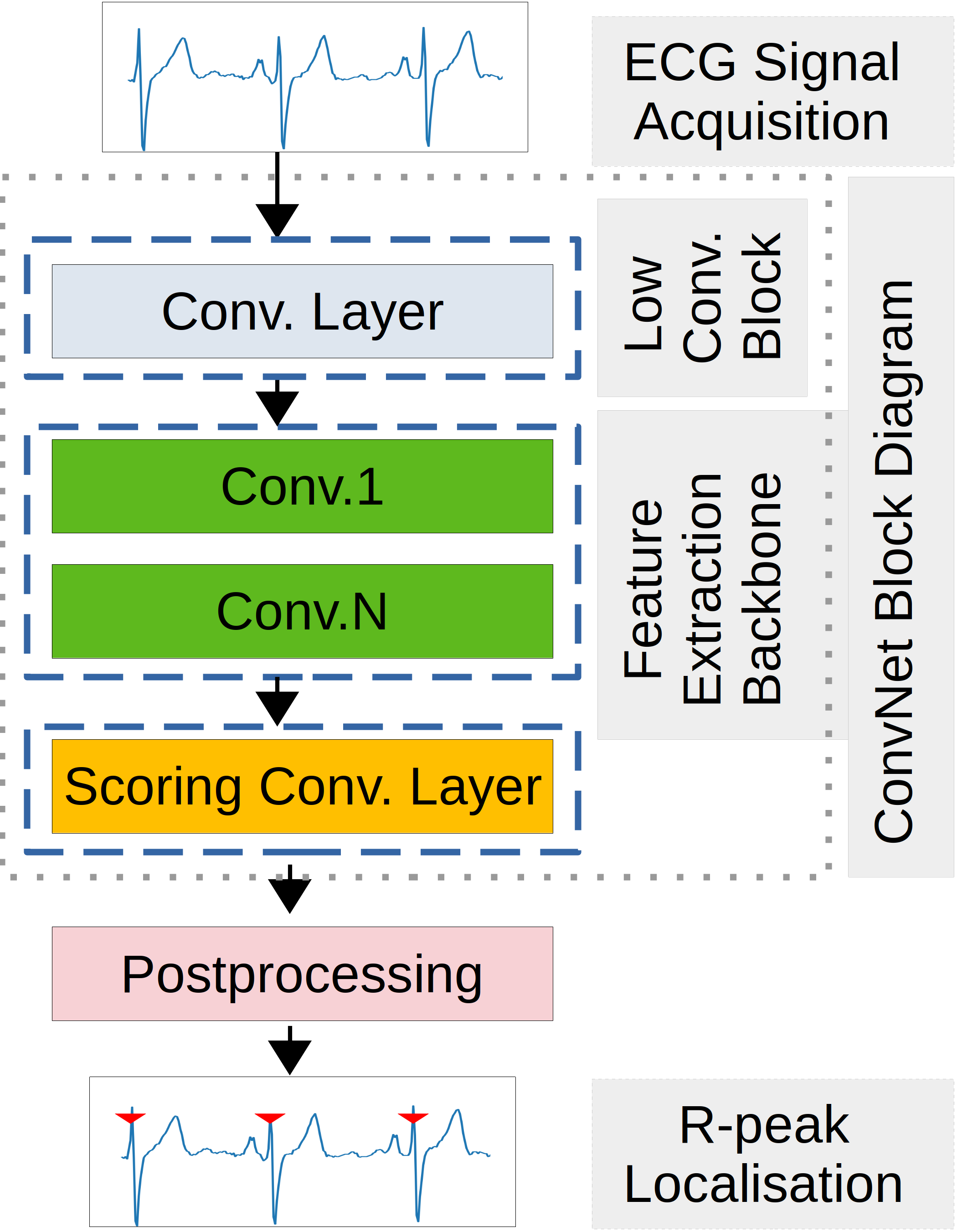}
  \end{subfigure}
  \caption{
  The schematic block-diagram of convolutional neural network used for R-peak localisation. 
  }
  \label{fig:network}
\end{figure}
The block-diagram of the network (Figure \ref{fig:network}) consists of a low-convolution-block, transforming raw-input into an intermediate representation, followed by a feature-extraction-block, consisting of one or more convolution-layer(s), and finally, a scoring-layer that projects into binary decision-planes to decide sample-wise predictions.
The convolution kernel hyper-parameter is optimised by adopting a domain-specific heuristic that the average duration of QRS-complex is 60 milliseconds and a 44 milliseconds equivalent (which is 5 samples for 100Hz signal) kernel should be long-enough to capture the most QRS features, following a study of Guo et al.~\cite{guo_inter-patient_2019}.
The CNN model was implemented in Python 3.7 using PyTorch 1.8.1 API.


\subsection{Model Training and Testing}
The MIT-BIH-Arrhythmia dataset is comparatively a less noisy dataset and was used for the model training to learn signal morphology, while the other datasets for cross-database testing.
The baseline-convnet was optimised by an internal subject-wise five-fold cross-validation of the MIT-BIH-Arrhythmia dataset, where subject-wise four-fold was used for training and the rest-fold subjects were used to validate the model and stop the training when suitable.
The five-fold approach yields five such optimised models, which were then used for cross-database testing, producing five validation scores per validation-dataset and finally, the scores were averaged.

An early-stopping mechanism was used in model-training, instead of using a fixed number of epochs, to early detect the convergence.
At the end of each epoch, the trained model was evaluated using one of the folds (internal five-fold approach, as mentioned above) and if the validation-loss does not improve for seven consecutive epochs (a.k.a. early-stopping patience), the training is stopped, while the maximum number of epochs was set to 50.
In most of the cases, it was found that the training process was terminated well before reaching the maximum epoch limit.
An initial learning rate was set to 0.01 to facilitate eager learning and a learning-rate-scheduler was used to decrease the rate by a factor of 0.1 of patience value set to 5, meaning that if the validation-loss does not decrease for five-consecutive epochs, the rate is decreased.
A cross-entropy loss was found suitable, possibly due to it's usage of natural logarithm that takes into account the difference between the actual label and predicted outcomes in a more granular way.
To back-propagate the calculated loss, the Adam optimiser was adopted for faster convergence \cite{kingma2014adam}.


\subsection{Post-processing Algorithm}
The post-processing tries to refine the CNN model prediction-stream just before the R-peaks are localised.
The post-processing is separated into three levels - minimal (Algorithm~\ref{algo:postprocessing_minimal}), moderate (Algorithm~\ref{algo:postprocessing_moderate}), and advanced level (Algorithm~\ref{algo:postprocessing_advanced}), based on the level of domain-specific knowledge required.
\begin{algorithm}[h]
    \caption{Salt and Pepper filter (Minimal post-processing).} \label{algo:postprocessing_minimal}
    \DontPrintSemicolon
    \KwData{
        \begin{itemize}
            \item PRED\_STREAM: Binary prediction stream (CNN model's output.)
            \item ONES\_PATTERNS: sequence of 1s with isolated 0s (11011, 1101).
            \item ZEROS\_PATTERNS: sequence of 0s with isolated 1s (00100, 0010).
        \end{itemize}}
    \KwResult{Refined binary prediction stream.}
    \SetKwInput{Kw}{Minimal: Salt and Pepper filter} \Kw{}
    \SetKwBlock{Begin}{begin}{end} \Begin{
      \nl \emph{Scan for ONES\_PATTERNS:} Scan binary stream to match ones patterns in sequence and replace with all ones of corresponding pattern length.\;
      \nl \emph{Scan for ZEROS\_PATTERNS:} Scan binary stream to match zeros patterns in sequence and replace with all zeros of corresponding pattern length.\;
      \nl \emph{Repeat above 2 steps:} Repeat scan for ONES\_PATTERNS and ZEROS\_PATTERNS until no match found.\;}
\end{algorithm}
\begin{algorithm}[h]
    \caption{Remove less confident QRS candidates (Moderate post-processing).} \label{algo:postprocessing_moderate}
    \DontPrintSemicolon
    \KwData{
        \begin{itemize}
            \item PRED\_STREAM: Binary prediction stream (CNN model's output.)
            \item DOMAIN\_KNOWLEDGE: Valid QRS extent should be at least 64 milli-seconds. Note that 64 milli-seconds of 100Hz signal yields 6 samples.
        \end{itemize}}
    \KwResult{Refined binary prediction stream.}
    \SetKwInput{Kw}{Moderate: Remove less confident QRS candidates} \Kw{}
    \SetKwBlock{Begin}{begin}{end} \Begin{
        \nl \emph{Do Salt and Pepper filtering (Algorithm~\ref{algo:postprocessing_minimal}).}\;
        \nl \emph{Calculate confidence score:} Scan all QRS candidates (occurrence of consecutive ones) in the binary prediction stream, where each candidate's confidence is the number of consecutive ones.\;
        \nl \emph{Filter out less confidants:} Scan all QRS candidates in the binary prediction stream and remove them with confidence score less than 6.\;}
\end{algorithm}
\begin{algorithm}[h]
    \caption{Remove short-distant QRS neighbors (Advanced post-processing).} \label{algo:postprocessing_advanced}
    \DontPrintSemicolon
    \KwData{
        \begin{itemize}
            \item PRED\_STREAM: Binary prediction stream (CNN model's output.)
            \item DOMAIN\_KNOWLEDGE: Two consecutive QRS nodes should be at least 200 milli-seconds (20 samples for 100Hz) distant (minimum R-R distance).
        \end{itemize}}
    \KwResult{Refined binary prediction stream.}
    \SetKwInput{Kw}{Advanced: Remove short-distant QRS neighbors} \Kw{}
    \SetKwBlock{Begin}{begin}{end} \Begin{
        \nl \emph{Remove less confident QRS candidates (Algorithm~\ref{algo:postprocessing_moderate}).}\;
        \nl \emph{Calculate confidence score:} See Algorithm~\ref{algo:postprocessing_moderate}\;
        \nl \emph{Calculate R-R interval:} Scan the QRS candidates and calculate the distance between two neighboring nodes.\;
        \nl \emph{Filter candidates with small R-R interval:} Scan the QRS candidates and for each node with R-R interval less than 20, remove either the candidate or its next neighbor whoever has comparatively a low confidence score.\;}
\end{algorithm}

The minimal post-processing (PP) is a basic Salt and Pepper filter which removes isolated ones (or zeros) recursively within a group of consecutive zeros (or ones).
Two patterns of consecutive ones (or zeros), with isolated zeros (or ones) inside, $11011$ and $1101$ (or $00100$ and $0010$), were searched in sequence for a match into the input binary stream, in order to replace them with all ones (or zeros) of corresponding pattern lengths (i.e. 5 and 4 ones). 
Matching the longer pattern $11011$ (or $00100$) before the shorter $1101$ (or $0010$) was found effective which seems maximised the lengths of consecutive ones (or zeros).
The moderate PP tries to filter out nodes with the confidence-score less than a threshold of 64 milliseconds equivalent number of samples~\cite{cai_qrs_2020} (approx. 6 samples for 100Hz signal) to remove QRS-like short lived artefacts. 
The advanced PP takes into account the R-R interval and filters out nodes whose R-R interval falls below a minimum threshold of 200 milliseconds equivalent number of samples (i.e. 20 samples for 100Hz signal worked well for our case, but Cai et al.~\cite{cai_qrs_2020} used 100 milliseconds).

\subsection {Validation score}
In this study, an F1-score was used as a validation score for proposed models.
F1-score is calculated as
\begin{equation}
F1 = 2 * \frac {PPV * Sensitivy} {PPV + Sensitivy}
\end{equation}

\section{Implementation}

\begin{figure}[h!]
  \centering
  \begin{subfigure}[b]{0.41\textwidth}
\includegraphics[width=\textwidth]{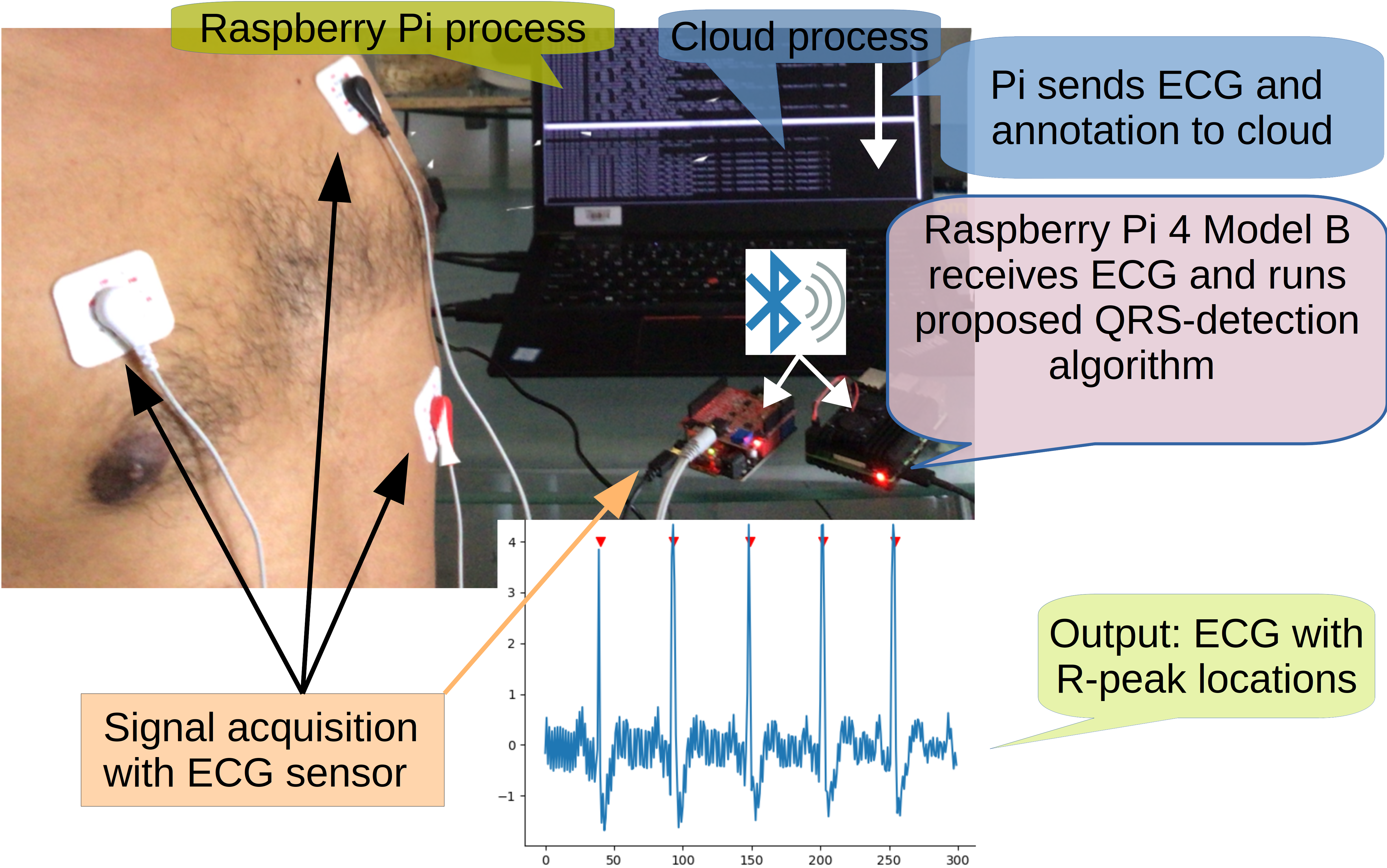}
  \end{subfigure}
  \caption{Implementation of the application framework (mentioned in section~\ref{app_framework}) pointing the hardware layer where the proposed QRS-detection solution fits into. 
}
  \label{fig:qrs_detect_process}
\end{figure}

An application framework that was shown in the Abstract, has been implemented in Raspberry Pi and shown in Figure~\ref{fig:qrs_detect_process}.
In this Raspberry Pi 4 model-B based implementation, ECG signal was received from the sensor over a Bluetooth low-energy interface, processed by our proposed set of QRS detection algorithms (set of configurations combining CNN model and post-processing complexity), and finally, sends the ECG signal, along with R-peak annotations, to the cloud using web interface.
Other low-resource devices were not considered for implementation in this study, rather a resource-configuration-based assumption has been discussed.

\section{RESULTS} \label{results}

\begin{figure}[h]
  \centering
  \begin{subfigure}[b]{0.31\textwidth}
    \includegraphics[width=\textwidth]{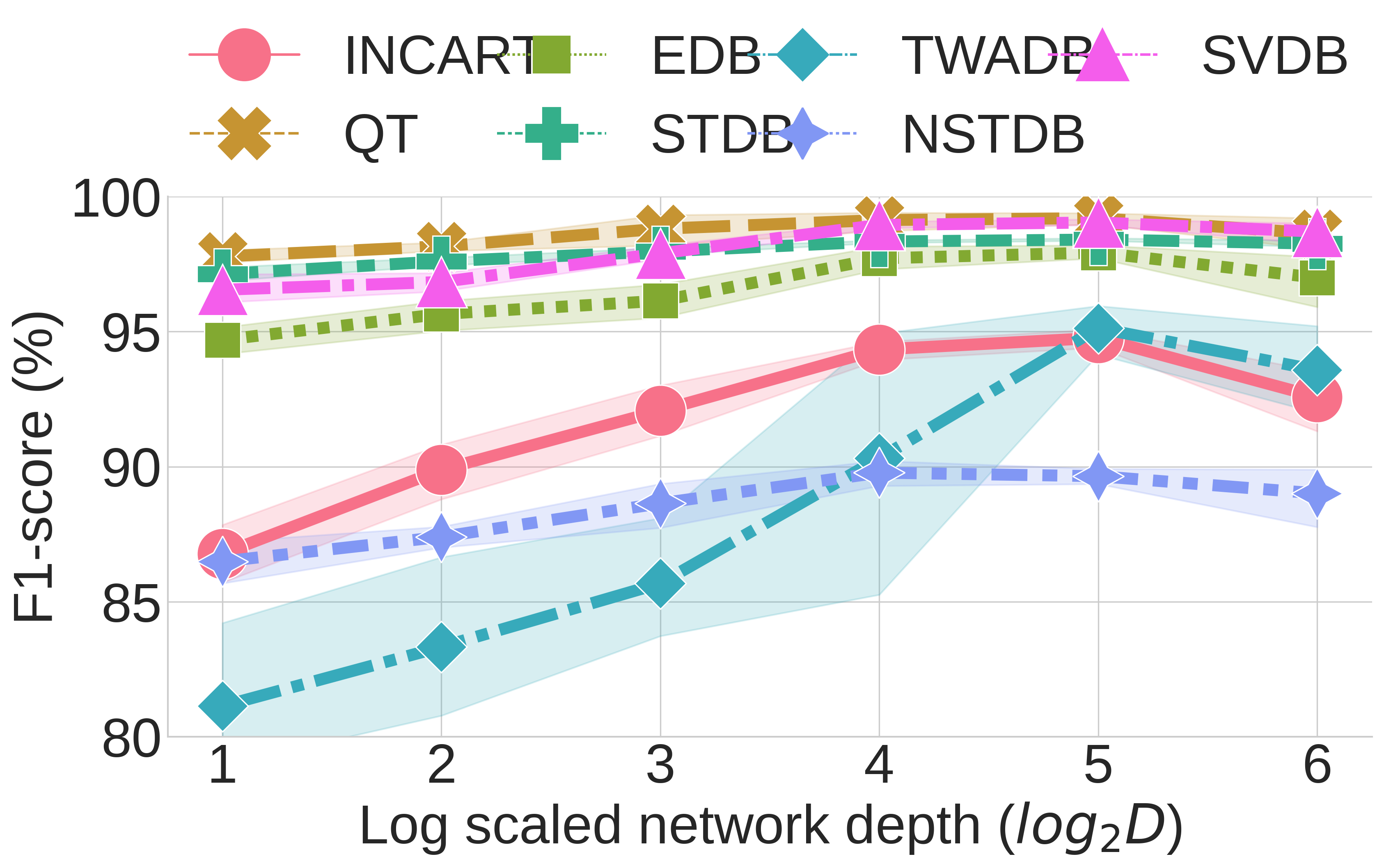}
    \caption{}
  \end{subfigure}
  \begin{subfigure}[b]{0.31\textwidth}
    \includegraphics[width=\textwidth]{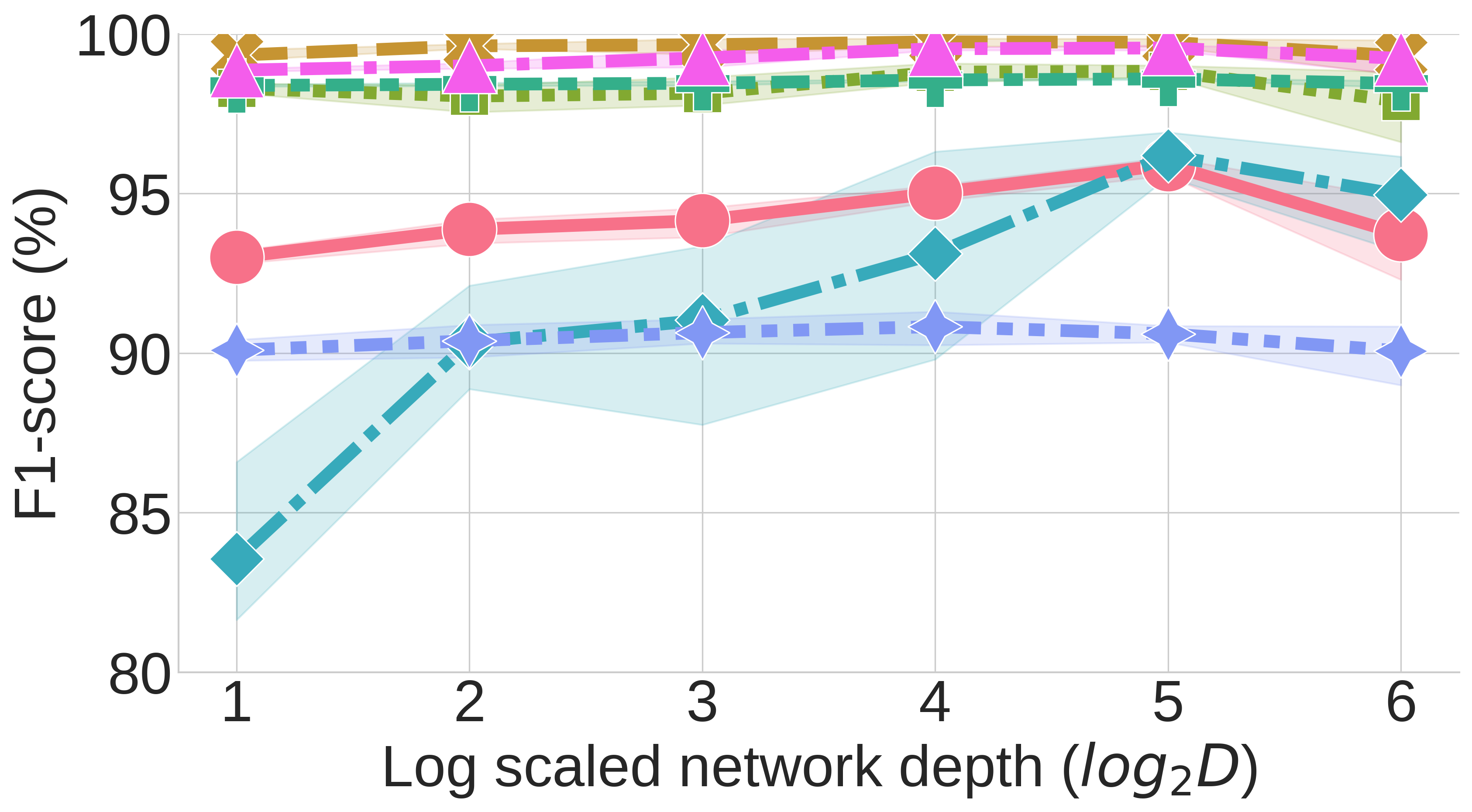}
    \caption{}
  \end{subfigure}
  \begin{subfigure}[b]{0.31\textwidth}
    \includegraphics[width=\textwidth]{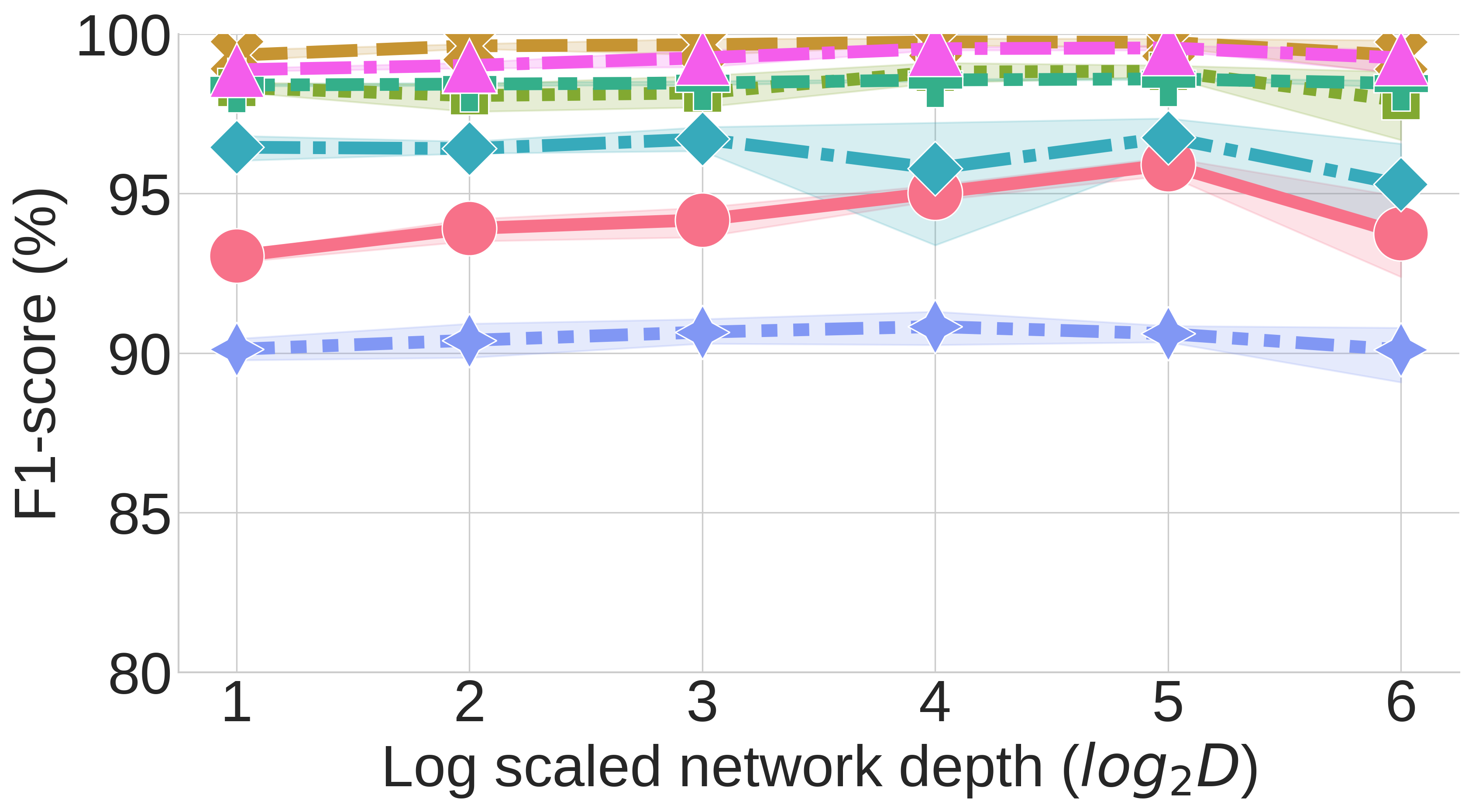}
    \caption{}
  \end{subfigure}
  \caption{F1-scores (\%) of various network depths (2, 4, 8, 16, 32, and 64-layers) with post-processing type (a) minimal (PP-1), (b) moderate (PP-2), and (c) advanced (PP-3). 
  }
  \label{fig:network_depths_vsteps}
\end{figure}

Figure~\ref{fig:network_depths_vsteps} shows the variation in F1-scores with increasing depths (scaled using $log_{2}D$) of the model using the post-processing (PP) methods - minimal, moderate, and advanced (indexed as PP 1-3, shown in Algorithm~\ref{algo:postprocessing_minimal},~\ref{algo:postprocessing_moderate}~\ref{algo:postprocessing_advanced}).
PP-1 F1-scores of 2-layer CNN (in Figure \ref{fig:network_depths_vsteps}-a) are the lowest and increase through 4 and 8-layer until 16-layer.
The F1-scores between 16 and 32-layer-depth marginally increases for INCART, QT, EDB, STDB, and SVDB, marginally decreases for the NSTDB, however, increases significantly for the TWADB (around 7\%).
Beyond 32-layer-depth, the F1-scores declined for all the datasets, where the INCART and TWADB decreased comparatively at a higher margin of around 3\%.
The NSTDB F1-score response varied around only 2\% (between 86.5\% and 89.8\%) across the depths, similar to a group of datasets (QT, EDB, STDB, SVDB), which varied $<$2\%, but the INCART and TWADB F1-scores varied around 8\% (between 87\% and 95\%) and 14\% (between 81\% and 95\%).


With the moderate and advanced PP (in Figure \ref{fig:network_depths_vsteps}-b,c), the F1-score growth variation flattens better compared to the minimal PP (in Figure~\ref{fig:network_depths_vsteps}-a) across the datasets. 
The advanced post-processed F1-score response is almost similar to the moderate post-processed scores across network-depths with an exception of TWADB, that the advanced PP shows significant improvement starting from 2-layer-deep network (improved by around 13\% margin, from 83\% in moderate to 96\% in advanced PP).

\begin{figure}[h]
  \centering
  \begin{subfigure}[b]{0.31\textwidth}
    \includegraphics[width=\textwidth]{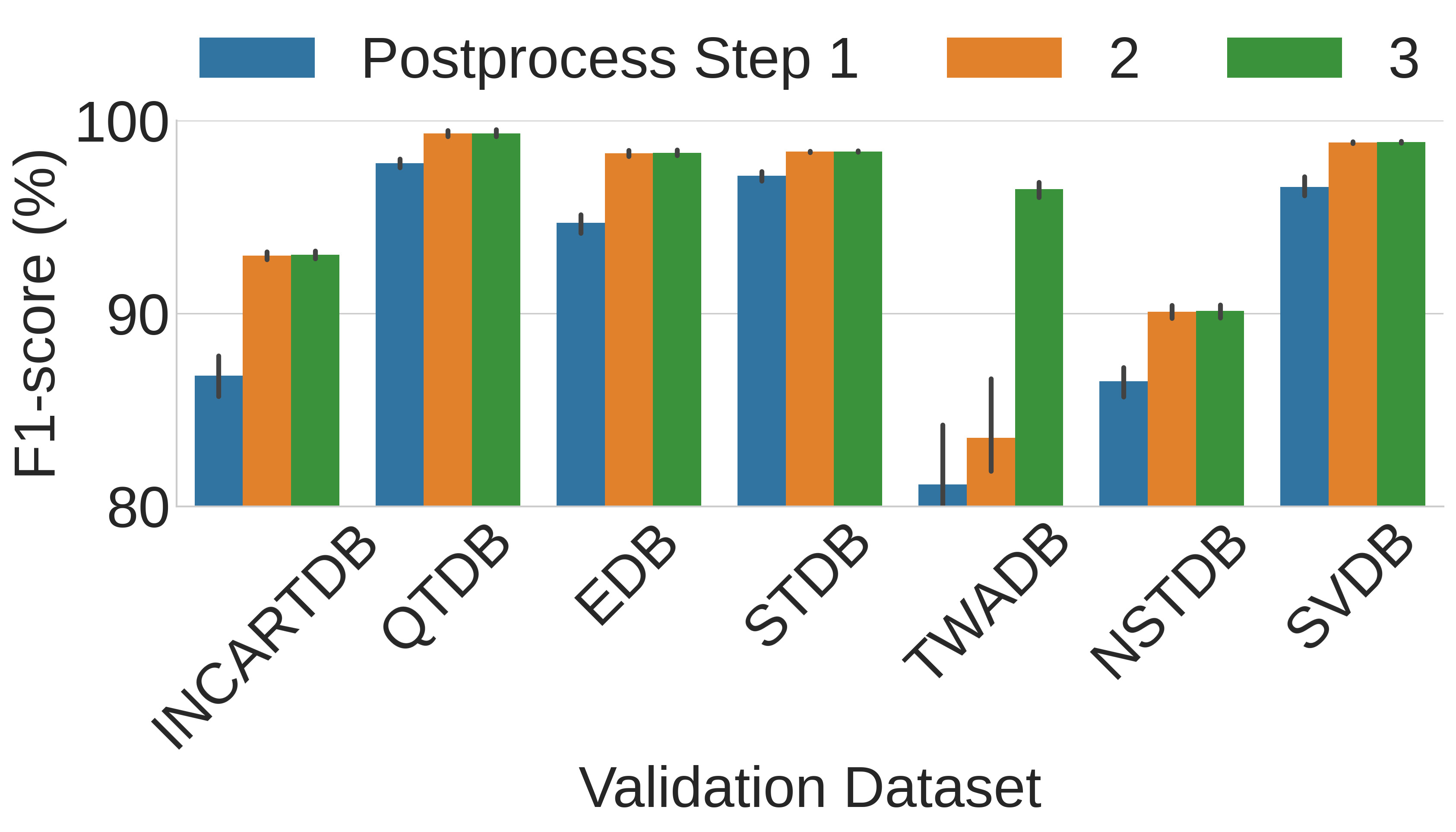}
    \caption{}
  \end{subfigure}
  \begin{subfigure}[b]{0.31\textwidth}
    \includegraphics[width=\textwidth]{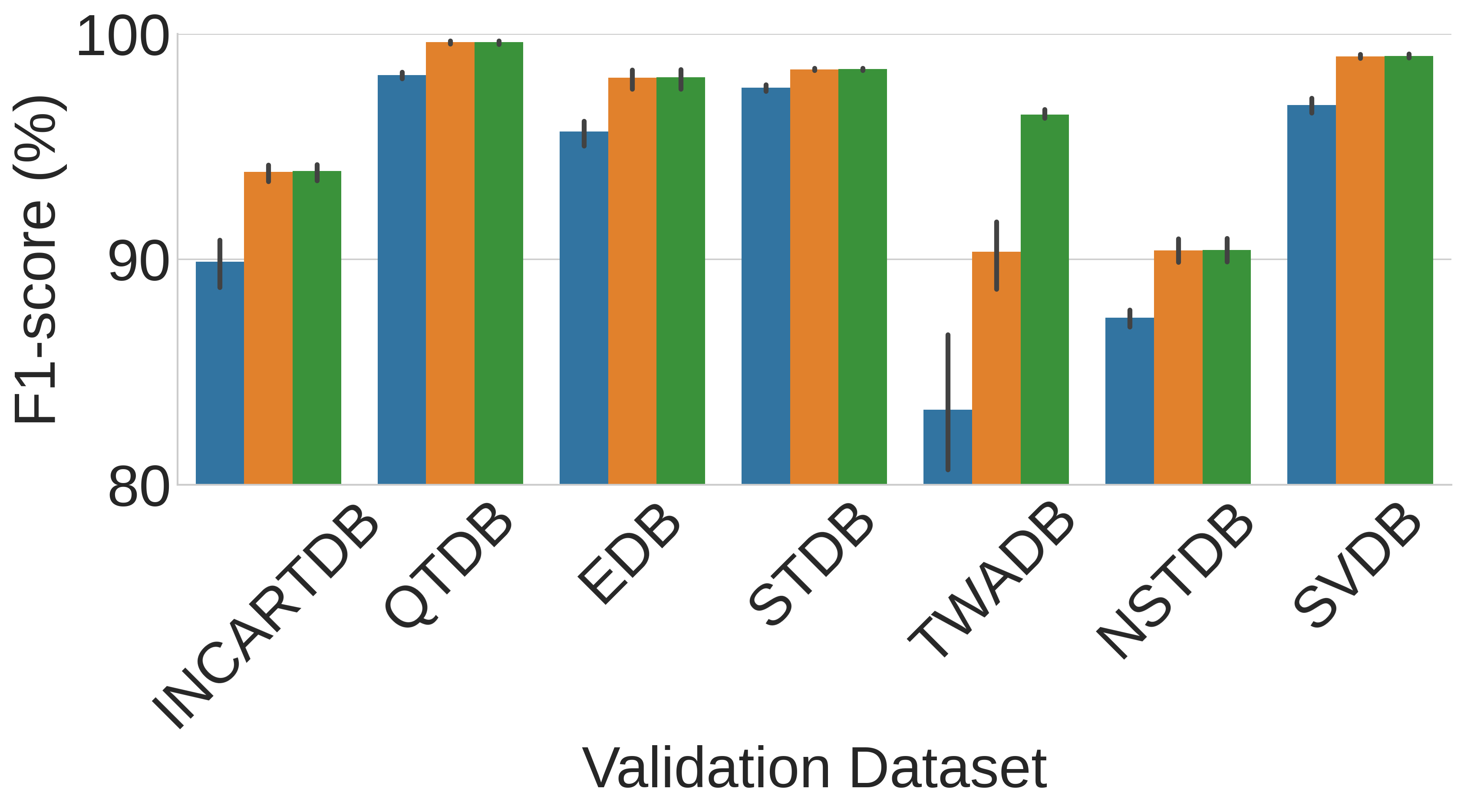}
    \caption{}
  \end{subfigure}
  \begin{subfigure}[b]{0.31\textwidth}
    \includegraphics[width=\textwidth]{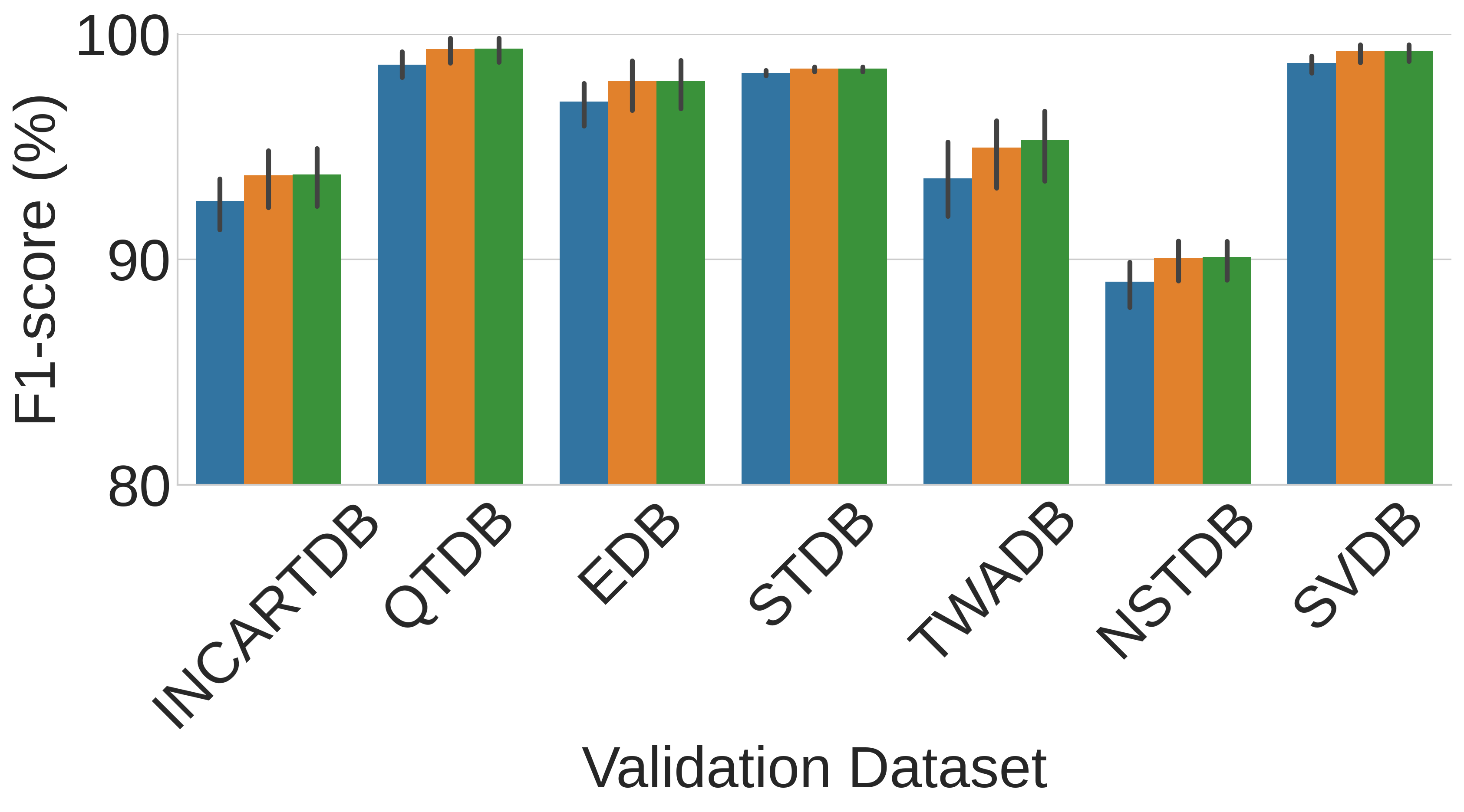}
    \caption{}
  \end{subfigure}
  \caption{
  Comparison of F1-scores of CNN models of depth (a) 2-layer, (b) 4-layer, 
  and (f) the deepest 64-layer, with three post-processing steps - minimal (step-1), moderate (step-2), and advanced (step-3). 
  }
  \label{fig:postprocessing_steps}
\end{figure}

Figure~\ref{fig:postprocessing_steps} shows three post-processing steps' relative importance across the validation datasets for networks with 2, 4, and 64 layer depths (intermediate depths not reported since the variation increases marginally, as well as, to keep it brief for better explainability).
\begin{figure*}[h]
  \centering
  \begin{subfigure}[b]{0.31\textwidth}
    \includegraphics[width=\textwidth]{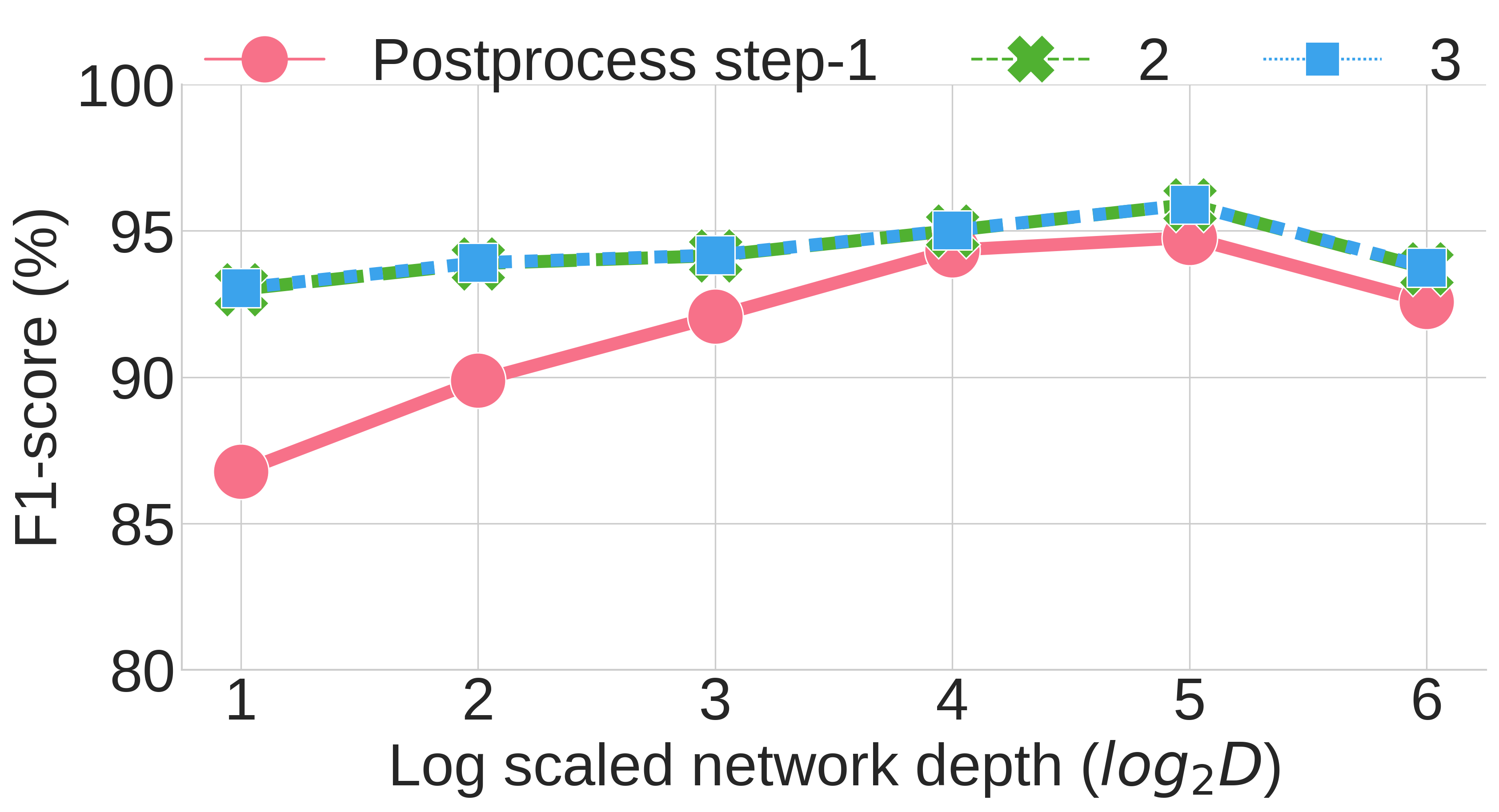}
    \caption{}
  \end{subfigure}
  \begin{subfigure}[b]{0.31\textwidth}
    \includegraphics[width=\textwidth]{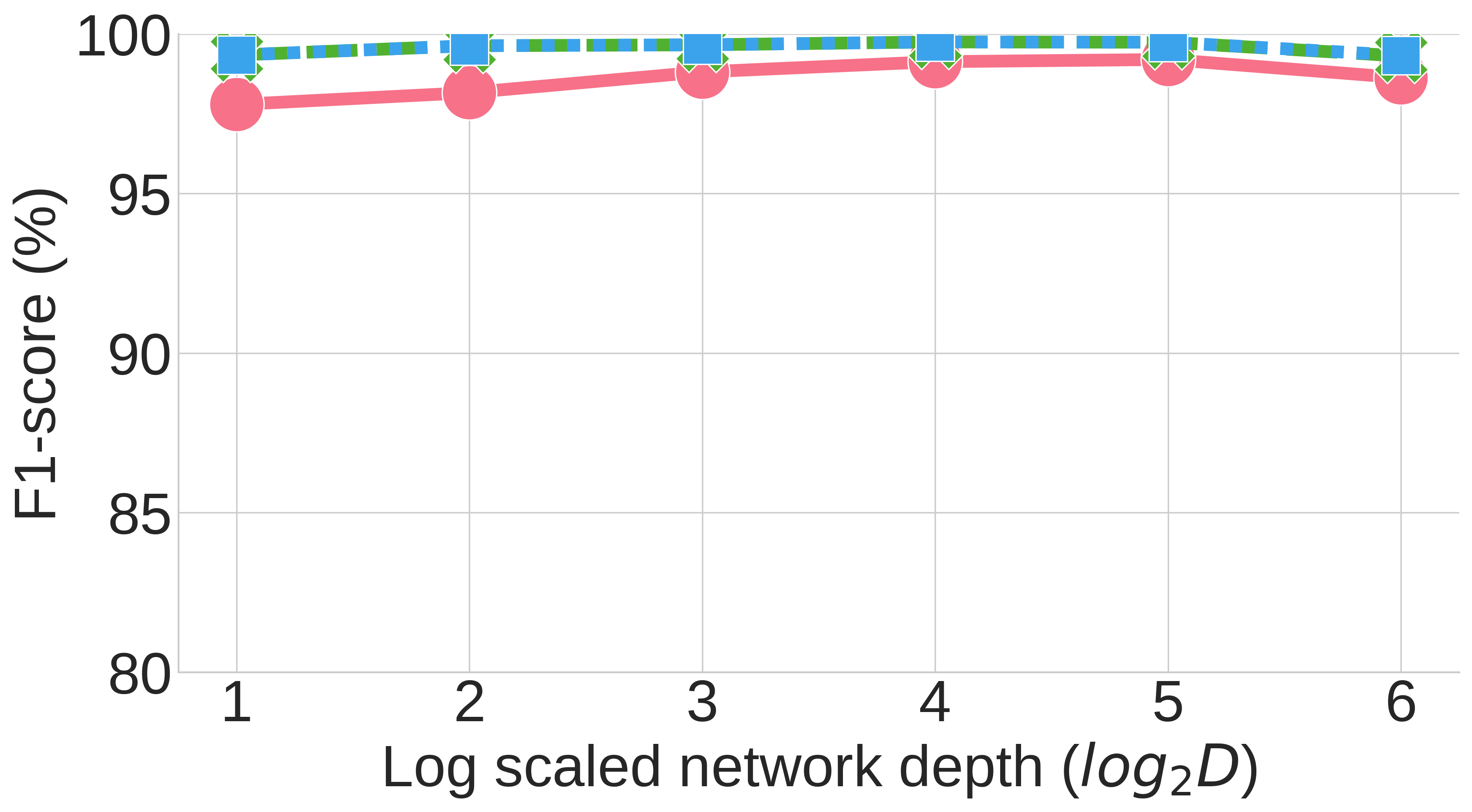}
    \caption{}
  \end{subfigure}
  \begin{subfigure}[b]{0.31\textwidth}
    \includegraphics[width=\textwidth]{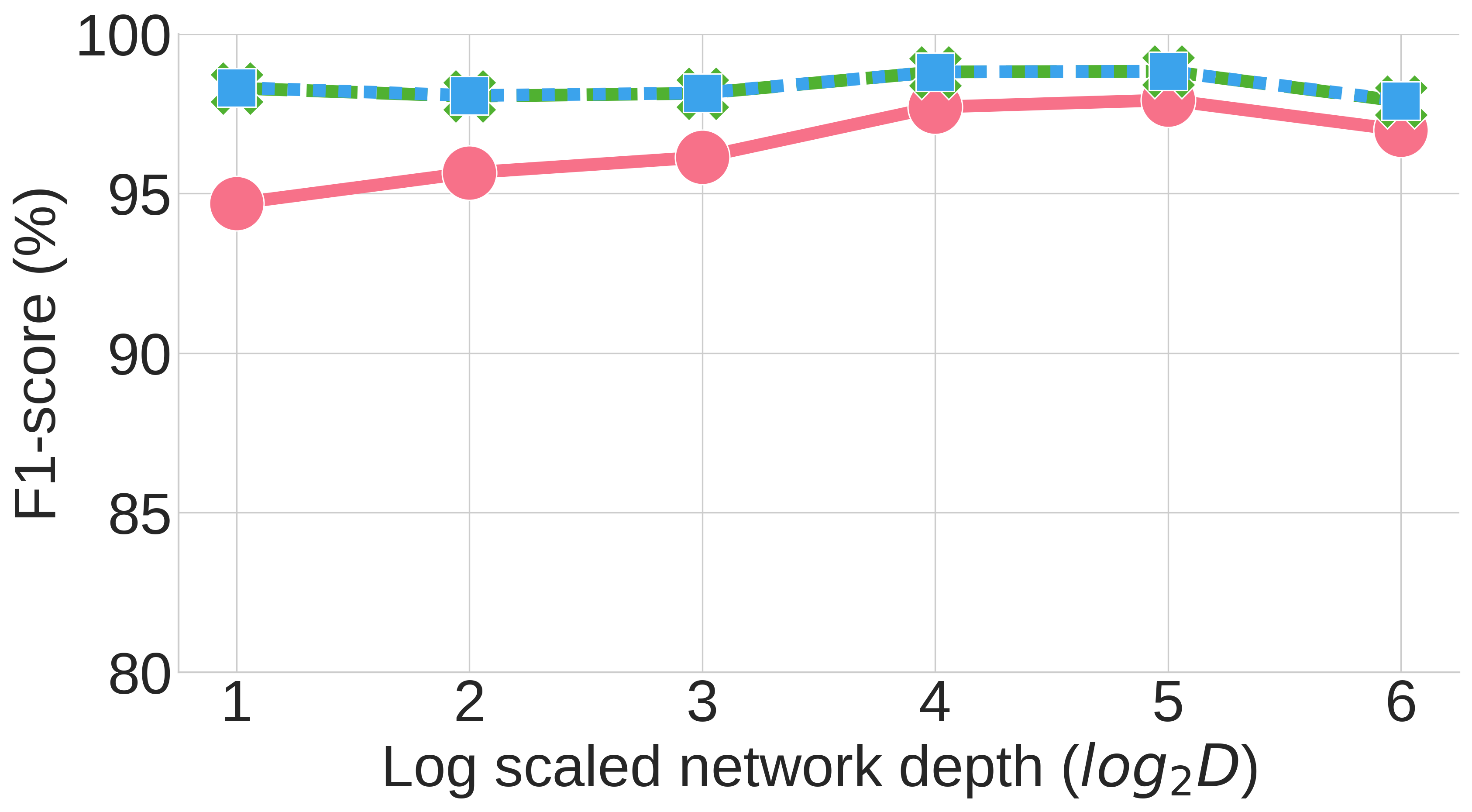}
    \caption{}
  \end{subfigure}
  \par\bigskip 
  \begin{subfigure}[b]{0.31\textwidth}
    \includegraphics[width=\textwidth]{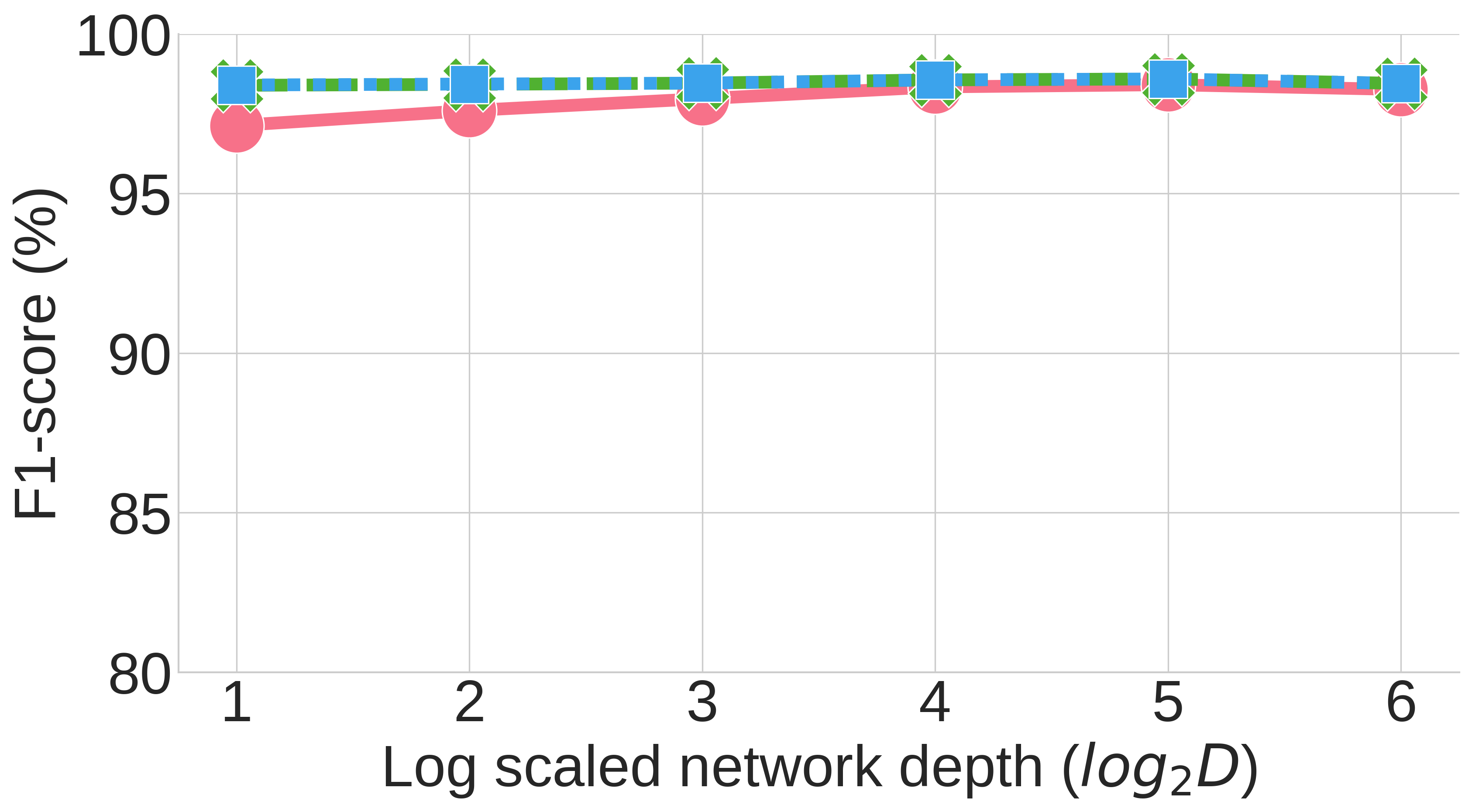}
    \caption{}
  \end{subfigure}
  \begin{subfigure}[b]{0.31\textwidth}
    \includegraphics[width=\textwidth]{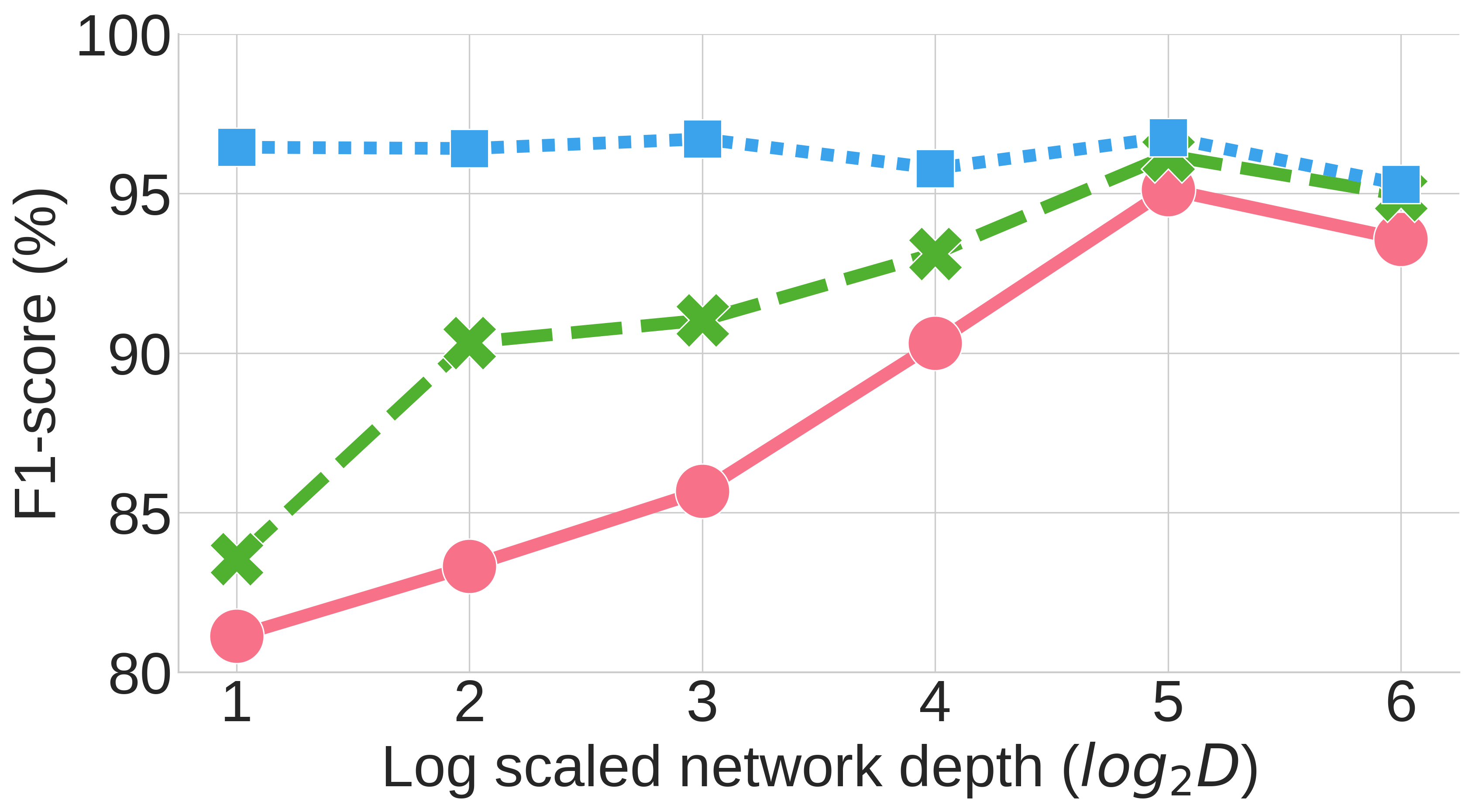}
    \caption{}
  \end{subfigure}
  \begin{subfigure}[b]{0.31\textwidth}
    \includegraphics[width=\textwidth]{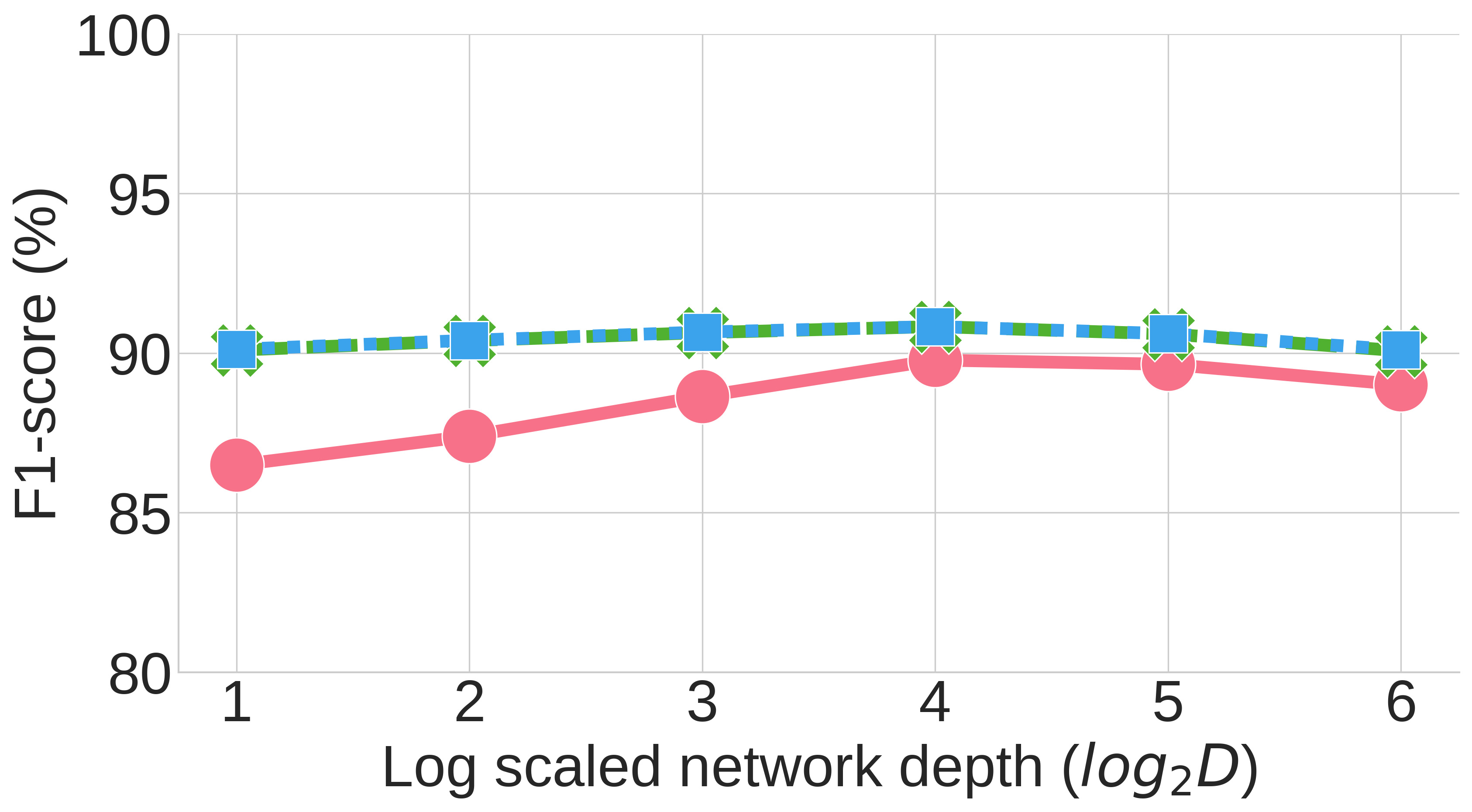}
    \caption{}
  \end{subfigure}
  \caption{
  Variation of F1-scores across network-depths of datasets, including (a) INCART, (b) QT, (c) EDB, (d) STDB, (e) TWADB, and (f) NSTDB, 
  for three post-processing methods (PP 1-3). The CNN model depths (2, 4, 8, 16, 32, and 64) are log scaled using $log_{2}D$.
  }
  \label{fig:score_depthwise_postprocess}
\end{figure*}
\begin{figure*}[h]
  \centering
  \begin{subfigure}[b]{0.31\textwidth}
    \includegraphics[width=\textwidth]{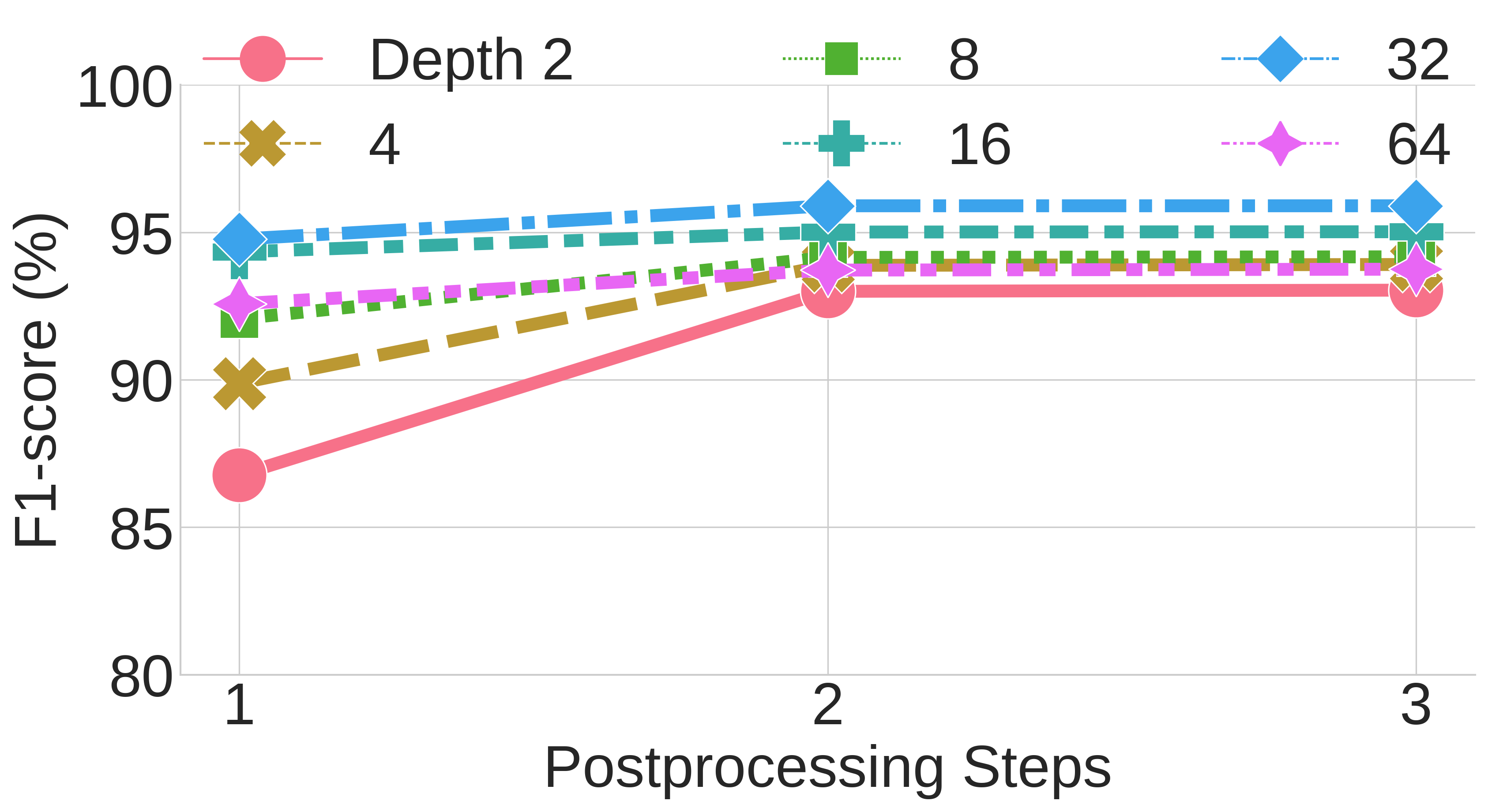}
    \caption{}
  \end{subfigure}
  \begin{subfigure}[b]{0.31\textwidth}
    \includegraphics[width=\textwidth]{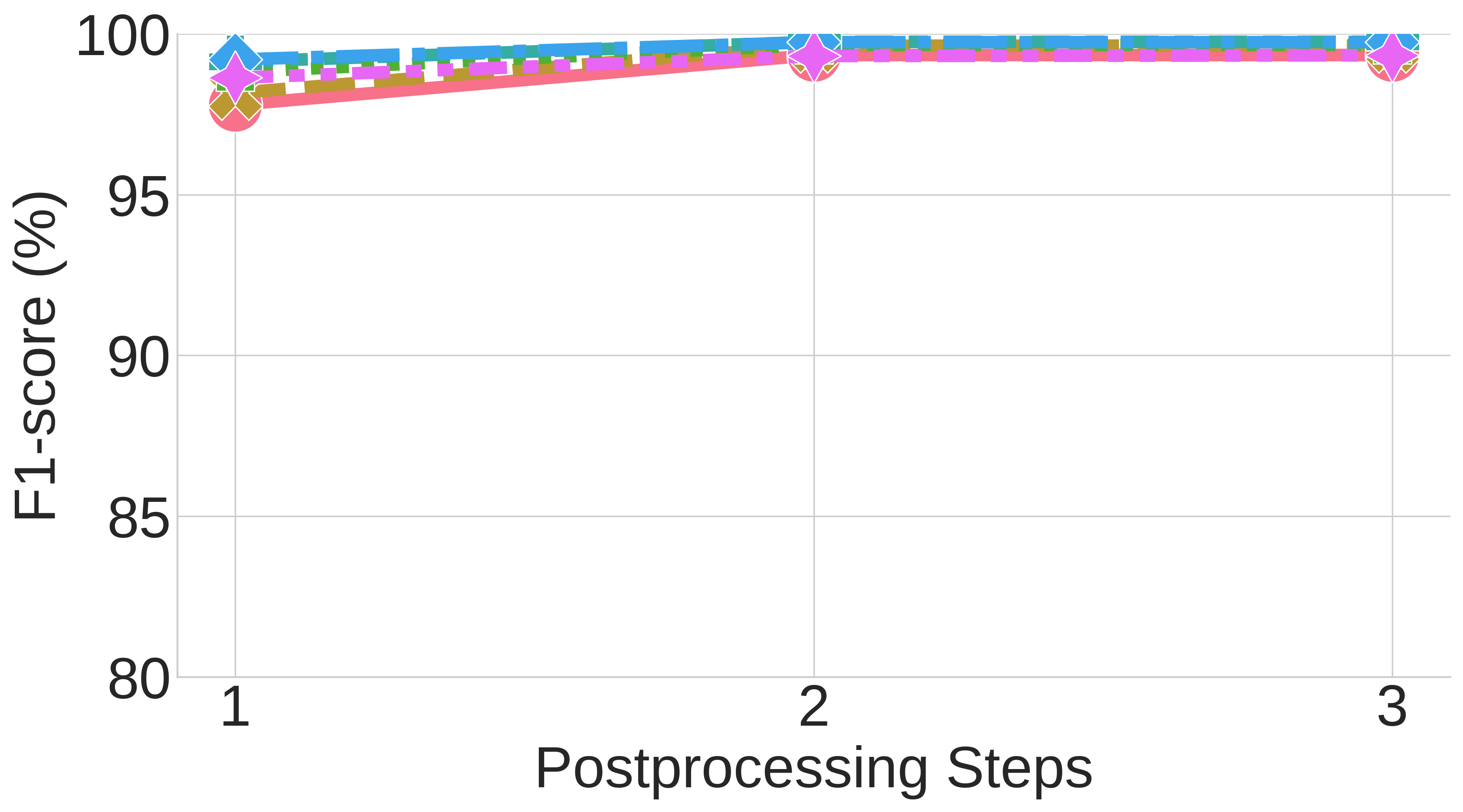}
    \caption{}
  \end{subfigure}
  \begin{subfigure}[b]{0.31\textwidth}
    \includegraphics[width=\textwidth]{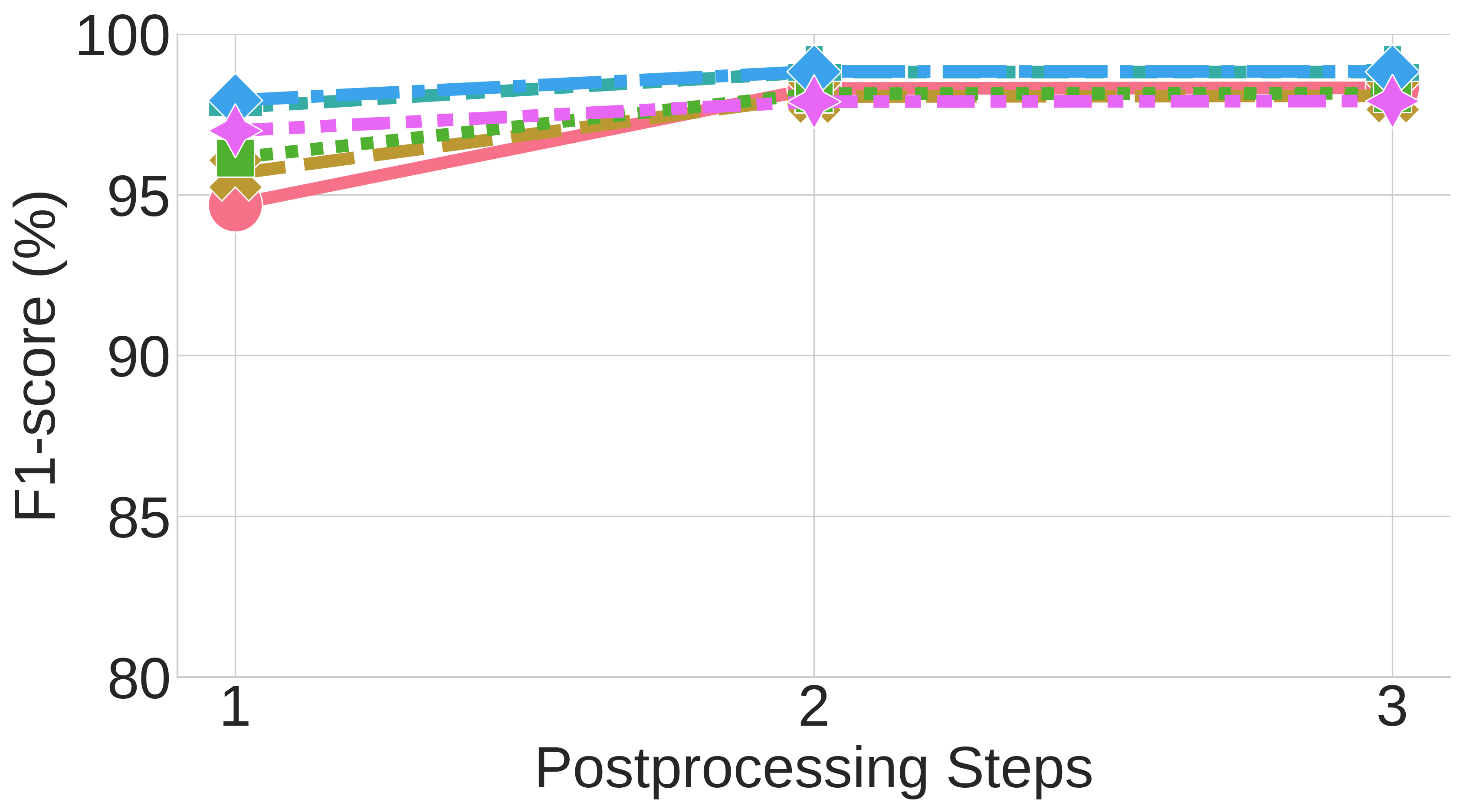}
    \caption{}
  \end{subfigure}
  \par\bigskip 
  \begin{subfigure}[b]{0.31\textwidth}
    \includegraphics[width=\textwidth]{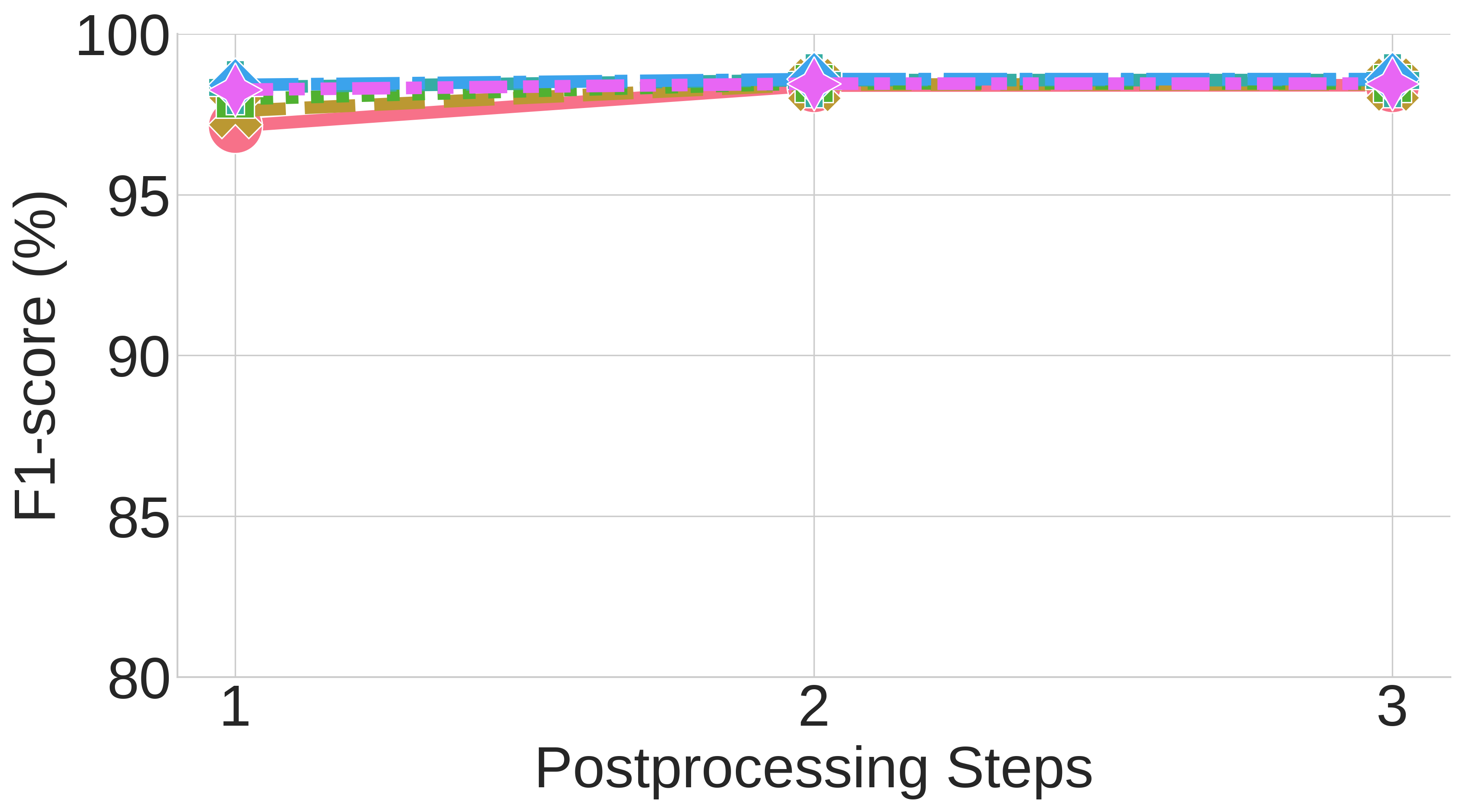}
    \caption{}
  \end{subfigure}
  \begin{subfigure}[b]{0.31\textwidth}
    \includegraphics[width=\textwidth]{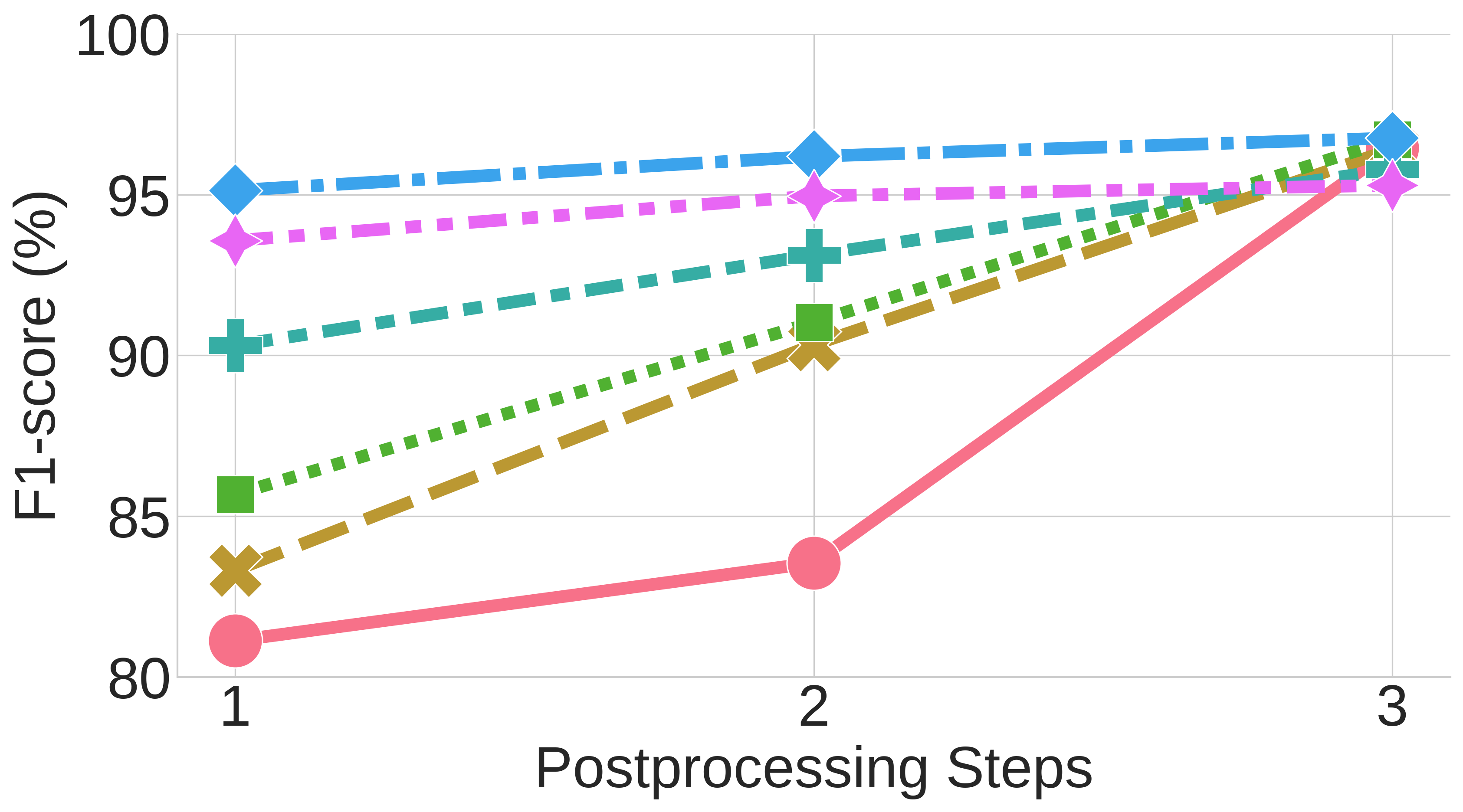}
    \caption{}
  \end{subfigure}
  \begin{subfigure}[b]{0.31\textwidth}
    \includegraphics[width=\textwidth]{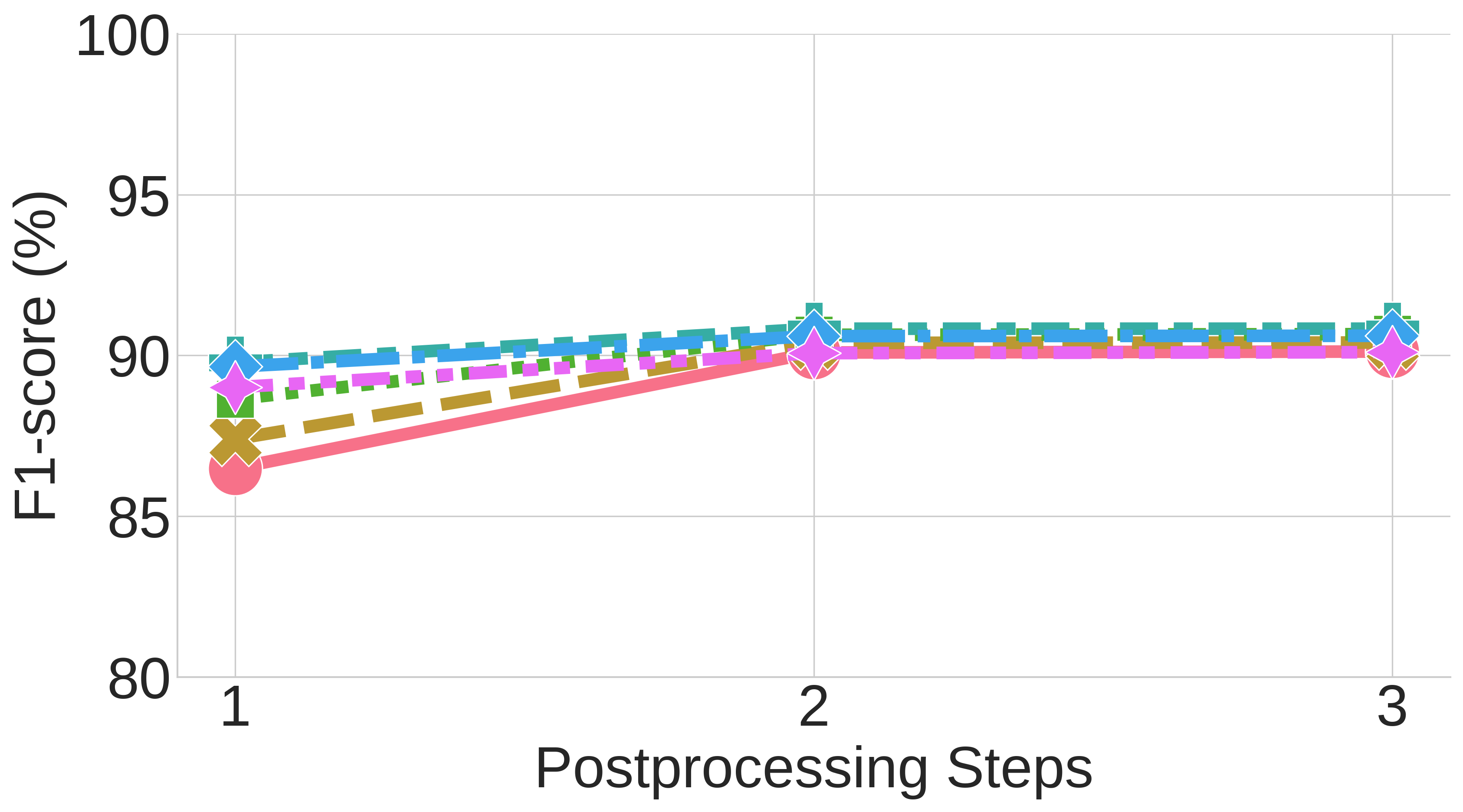}
    \caption{}
  \end{subfigure}
  \caption{
  Variation of F1-scores across post-processing levels (i.e. minimal, moderate, and advanced, indexed as PP 1-3), for network-depths (of 2, 4, 8, 16, 32, and 64-layer) of datasets, including, (a) INCART, (b) QT, (c) EDB, (d) STDB, (e) TWADB, and (f) NSTDB.
  }
  \label{fig:score_postprocesswise_depth}
\end{figure*}
Three post-processing methods (indexed as PP 1-3) are represented as blue, yellow, and green color-bar in order.
In Figure~\ref{fig:postprocessing_steps}-a, the PP-1 (blue-bar) shows the lowest F1-scores, compared to the other two PPs, while its lowest score is around 81\% for TWADB.
PP-2 (yellow-bar) is higher than PP-1 (blue-bar), for the 2-layer-depth (Figure~\ref{fig:postprocessing_steps}-a), where this margin is $<$5\% for all the datasets.
The PP-3, in the same Figure~\ref{fig:postprocessing_steps}-a, on the other hand, is superior to the PP-2 marginally for all, but the TWADB, for which the margin is around 15\%.
With 4-layer deep CNN (Figure~\ref{fig:postprocessing_steps}-b), the scores of PP-1 improved proportionately, but the TWADB and NSTDB marginally improved ($<$2\%).
The highest 64-layer deep CNN (Figure~\ref{fig:postprocessing_steps}-c) almost equalises the effects of the PP methods (PP 1-3) for the datasets, including the TWADB, where PP-1 is lower than PP-3 by only around 1\% margin.

The characteristic curves, in the Figure~\ref{fig:score_depthwise_postprocess}, depicts the score variation of different PPs w.r.t. the network-depths, for each validation dataset.
For almost all the datasets, the PP-2 and PP-3 are indistinguishable, except TWADB, for which all PPs tend to come closer at 32-layer-deep CNN (shown in Figure~\ref{fig:score_depthwise_postprocess}-e).
The PP-2 is marginally ($<$1\%) lower than PP-3 for datasets like QTDB, and STDB (in Figure~\ref{fig:score_depthwise_postprocess}-b, and d) across network depths, but this difference is in the range of 2-10\% for datasets like INCART, EDB, NSTDB, (in Figure~\ref{fig:score_depthwise_postprocess}-a, c, f) and $>$12\% for TWADB (in Figure~\ref{fig:score_depthwise_postprocess}-e).


The characteristic curves in Figure \ref{fig:score_postprocesswise_depth} shows the response of F1-scores for network-depths w.r.t. PPs, across the validation datasets.
The shallow 2-layer CNN, with PP-1, yields the lowest F1-scores across the datasets, but with the PP-2, the score improves (with margin in the range around 1-6\%).
The PP-3, on the other hand, always achieves marginal F1-score improvement compared to the moderate post-processing, with the exception of TWADB (Figure \ref{fig:score_postprocesswise_depth}-e), that improved with around 13\% margin.
Unlike the INCART and TWADB, the 8, 16, and 32-layer-deep networks are almost indistinguishable for the rest validation-datasets across PPs.

CNN models were trained with large amount of data in computing hardware (GPU server with Nvidia Tesla Volta V100-SXM2-32GB), where the amount of additional time, required for each training epoch for increasing depths of CNN models (shown in Table~\ref{tab:cnn_param}), increases exponentially.
\begin{table}[h]
    \centering
    \caption{Model statistics and post-processing (PP) complexity for various depths of CNN models, used in this study.}
    \begin{tabular}{ccccccc}
      \hline
      \emph{No. of Layers:} & 2 & 4 & 8 & 16 & 32 & 64 \\
      \hline
      \emph{\makecell{CNN \\no. of param:}} & 6244 & 12100 & 23812 & 47236 & 94084 & 187780 \\
      \hline
      \emph{\makecell{CNN memory\\in kB:}} & 34 & 61 & 117 & 227 & 448 & 891 \\
      \hline
      \emph{\makecell{Avg. train time\\/mini-batch}} & 26s & 33s & 46s & 70s & 120s & 220s \\
      \hline
      \emph{\makecell{Avg. test time\\for INCART}} & $<$8s & $<$8s & $<$8s & $<$8s & $<$8s & $<$8s \\
      \hline
      \emph{\makecell{PP Complexity\\minimal, \\moderate, \\or advanced:}} & $\mathcal{O}(n)$
 & $\mathcal{O}(n)$ & $\mathcal{O}(n)$ & $\mathcal{O}(n)$ & $\mathcal{O}(n)$ & $\mathcal{O}(n)$ \\
    \end{tabular}
    \label{tab:cnn_param}
\end{table}
The table also shows the number of CNN model parameters, and memory requirement (in kB), which also grow exponentially with the increase of model depths.
The models requires more memory, due to the number of parameters, for higher depths.
The validation or test time, on the other hand, is constant across network depths, as shown for INCART dataset, run in Raspberry Pi 4. 
Each post-processing method scans the output prediction-stream sequentially and updates the individual bits of the stream as required, thus, their run-time complexity is $\mathcal{O}(n)$.

\section{DISCUSSION} \label{discussion}

IoMT solutions often seek traditional filter-based algorithms for embedded device-based resource constrained environments and reluctant to use deep-learning-based algorithms due to their high resource requirement, in spite of their strengths including noise resistance, automated feature extraction and learning. 
In a sensor-based end-to-end solution, a shallow CNN model has the capacity to offer better performance by leveraging other core components, such as post-processing.
To do that, in this study, the complexity of CNN models (only the depth parameter varied, since it solely dominates a model's learning capacity) were incrementally increased and combined with a set of post-processing.
A use case of  IoMT based ECG signal monitoring was chosen, where the post-processing was segregated based on an increasing use of domain-knowledge including minimal PP-1 (basic salt \& pepper filter), moderate PP-2 (minimum QRS extent 64 milli-seconds), and advanced PP-3 (minimum R-R distance to be 200 milli-seconds).
Results of this study shows that it is possible to design an optimal DL solution with known target performance characteristics and resource (computing capacity and memory) constraints.

A combination of a shallow 2-layer CNN and PP-1 seems capable to achieve more than 85\% F1-score for almost all the datasets (except TWADB, that scores around 81\%).
If this is considered as a basic configuration, a superior performance can be achieved by altering the post-processing only.
For example, with PP-2, and PP-3, the score of a shallow 2-layer CNN improves by 3\% and 15\% w.r.t. PP-1.
Different CNN model architecture and complexity were used in the QRS detection literature~\cite{cai_qrs_2020,liu_octave_2020,laitala2020robust,jia_high_2019}, however, the use of a shallow 2-layer CNN model is scarce, where the emphasis was put on post-processing.
The strength of post-processing can be leveraged to process a primitive CNN model's output to achieve comparable performance across a broad range of datasets.
Such a configuration would be embedded device friendly, both in terms of CNN model (2-layer CNN has the lowest memory footprint), and post-processing (linear run-time complexity).

The quality of CNN model's prediction-stream improves with the increase of its complexity.
This can be understood by observing the improvement of PP-1 F1-scores for CNNs with increasing depths (characteristic curve, shown in Figure~\ref{fig:score_depthwise_postprocess}).
The PP-1 is a basic Salt \& Pepper filter, which was found to reach almost the highest level scores of PP-3 for CNN model of depth-8 (QT, and STDB, in Figure~\ref{fig:score_depthwise_postprocess}-b,d), depth-16 (INCART, EDB, and NSTDB, in Figure~\ref{fig:score_depthwise_postprocess}-a,c,f), and depth-32 (TWADB, in Figure~\ref{fig:score_depthwise_postprocess}-e).
Deep CNN models (depth $>$8 layers) learn better from the training data~\cite{simonyan_very_2015}, yield better prediction-stream and reduces the effect of post-processing, although, each dataset has its own depth requirement to achieve the top score.
It is particularly important to observe that a shallow 2-layer CNN model may be difficult to train to learn a complex function~\cite{seide2011conversational,simonyan_very_2015} (i.e. if an input sample belongs to a QRS region or not), that requires PP-2 (for most datasets, in Figure~\ref{fig:score_depthwise_postprocess},\ref{fig:score_postprocesswise_depth}) or PP-3 (for all datasets) to perform better than PP-1, because it seems that the use of domain-knowledge-based post-processing is complementing the shortcoming of the capacity of a shallow model.
The selection of PP-2 or PP-3 (instead of PP-1) with deep CNN models (depth $>$8 layers) was found to be a better choice to achieve the top performance for most of the datasets (except TWADB, for which PP-3 is required, instead of PP-2).
This requirement, however, may not comply with a target resource-constrained device, which has memory or computation limitation, thus, requires a shallow CNN model.

For a CNN model to have greater generalisability, it is required to perform almost at the same level for a broad range of test ECG data.
The characteristic curves of datasets, w.r.t. PPs, or CNN depths (in Figure~\ref{fig:score_depthwise_postprocess},~\ref{fig:score_postprocesswise_depth}), show the peculiarity of some of them (i.e. INCART, and TWADB), which achieves better performance with a resource-constrained configuration (shallow 2-layer CNN with PP-3), but requires a deep CNN model (8 or 16-layer CNN with PP-3) to achieve the peak performance.
A deep CNN model seems trying to understand subtle patterns which enables it to produce better prediction-stream for a wide range of test records (with greater inter and intra-patient variance).
As one can assume, the minimum configuration (shallow 2-layer CNN with PP-3) may not guarantee a similar level of performance for a test ECG record if it is much different than the training set and the shallow CNN likely be failing to learn the general features during training.
Data diversity is one of the challenges in designing a universal QRS-detector~\cite{habib_impact_2019}, and there are studies in the literature, which cross-validates the model with fractions of a single dataset or a few datasets.
Having a deep CNN model to generate better prediction-stream seems beneficial for a clinical context where the performance is the prime requirement.

A suitable configuration can be derived for a target resource-constraint device by matching its memory and computation capacity with those of the CNN models.
These observations can be summarised as below points, considering a simplified assumption that the validation datasets represent a universal set of datasets -
\begin{itemize}
    \item Target device has \emph{extreme low} memory and computing capacity: a combination of 2-layer CNN and PP-3 would yield the best performance.
    \item Target device has \emph{low} memory and computing capacity: a combination of 8-layer CNN and PP-3 yields the top performance (for all but the TWADB).
    \item Target device has \emph{sufficient} resources: a combination of 32-layer CNN and PP-2 or PP-3 yields the top performance for all the datasets.
\end{itemize}

\section{CONCLUSION} \label{conclusion}

Deep-learning models often are not a default choice of method in an IoMT end-to-end applications which require such models to deploy in resource constrained environment.
Classical filter-based approaches are commonly encountered in such context due to their small memory and computation footprint.
In this study, we explored ways for optimising DL solutions for improving their deployability in resource (computation capacity and memory) constrained devices. 
In a QRS-detection case study, it was shown that by effectively combining CNN models and post-processing by varying their complexities could facilitate to yielding a set of configurations suitable for a range of target resource constrained environments.
It was found that a narrow 2-layer CNN can achieve comparable performance with a suitable post-processing, targeting a low-resource environment, however, $>$8-layer CNN would be required to achieve the top scores, which may be suitable for comparatively rich target environments.
Results of this study pave ways to design DL solution for known resource constraint and target performance characteristics.
Finally, this study contributes in improving deploy-ability of DL solutions in resource constrained devices.

\section{Future Work}

CNN model's complexity optimisation to yielding a compositional solution to test on real-life data for a given task (i.e. QRS-detection and R-peak localisation) to monitor performance and power-consumption of resource constrained environment could be a promising future direction.


\section*{Acknowledgment}

This research was undertaken with the assistance of resources and services from the National Computational Infrastructure (NCI), which is supported by the Australian Government.

\bibliographystyle{IEEEtran}
\bibliography{zotero}

\end{document}


\begin{table*}[ht!]
  \centering
  \caption{Cross-database validation PPV (\%), Sensitivity (\%), and F1-scores (\%) of 2, 4, 8, 16, 32, and 64-layer deep baseline-convnet with minimal post-processing. For brevity, the scores were rounded to whole numbers.}
    \begin{tabular}{ccccccccccccccccccc}
    \hline
    Depth & \multicolumn{3}{c|}{2} & \multicolumn{3}{c|}{4} & \multicolumn{3}{c|}{8} & \multicolumn{3}{c|}{16} & \multicolumn{3}{c|}{32} & \multicolumn{3}{c}{64} \\
    \hline
    \emph{Validation} & \emph{PPV} & \emph{Se} & \emph{F1} & \emph{PPV} & \emph{Se} & \emph{F1} & \emph{PPV} & \emph{Se} & \emph{F1} & \emph{PPV} & \emph{Se} & \emph{F1} & \emph{PPV} & \emph{Se} & \emph{F1} & \emph{PPV} & \emph{Se} & \emph{F1} \\
    \hline
    INCART & 79 & 97 & 87 & 84 & 96 & 90 & 89 & 96 & 92 & 94 & 95 & 94 & 93 & 97 & 95 & 91 & 94 & 93 \\ 
    QT & 96 & 100 & 98 & 97 & 100 & 98 & 98 & 100 & 99 & 98 & 100 & 99 & 99 & 100 & 99 & 98 & 100 & 99 \\ 
    EDB & 91 & 99 & 95 & 93 & 99 & 96 & 94 & 99 & 96 & 96 & 99 & 98 & 97 & 99 & 98 & 96 & 98 & 97 \\ 
    STDB & 97 & 98 & 97 & 98 & 98 & 98 & 98 & 98 & 98 & 99 & 98 & 98 & 99 & 98 & 98 & 99 & 98 & 98 \\ 
    TWADB & 69 & 98 & 81 & 73 & 98 & 83 & 76 & 98 & 86 & 85 & 96 & 90 & 93 & 97 & 95 & 91 & 96 & 94 \\ 
    NSTDB & 79 & 96 & 87 & 81 & 96 & 87 & 83 & 95 & 89 & 86 & 94 & 90 & 85 & 95 & 90 & 84 & 95 & 89 \\ 
    SVDB & 93 & 100 & 97 & 94 & 100 & 97 & 96 & 100 & 98 & 98 & 100 & 99 & 98 & 100 & 99 & 98 & 100 & 99 \\ 
    \hline
  \end{tabular}
  \label{tab:scores}
\end{table*}